\documentclass[lettersize,journal]{IEEEtran}
\usepackage{amsmath,amsfonts}
\usepackage{algorithmic}
\usepackage{algorithm}
\usepackage{array}
\usepackage[caption=false,font=normalsize,labelfont=sf,textfont=sf]{subfig}
\usepackage{textcomp}
\usepackage{stfloats}
\usepackage{url}
\usepackage{colortbl}
\usepackage{verbatim}
\usepackage{graphicx}
\usepackage{makecell}
\usepackage{cite}
\usepackage{caption}
\usepackage{subfloat}
\usepackage{bbding}
\usepackage{color}
\usepackage{multirow}
\usepackage[percent]{overpic}
\usepackage[pagebackref,breaklinks,colorlinks,urlcolor=black]{hyperref}
\usepackage{booktabs}
\usepackage[linewidth=1pt]{mdframed}
\usepackage{arydshln}
\usepackage{soul}
\usepackage[thicklines]{cancel}

\RequirePackage{xspace}

\makeatletter
\DeclareRobustCommand\onedot{\futurelet\@let@token\@onedot}
\def\@onedot{\ifx\@let@token.\else.\null\fi\xspace}

\def\eg{\emph{e.g}\onedot} 
\def\ie{\emph{i.e}\onedot} 
 
\def\etc{\emph{etc}\onedot}

\def\etal{\emph{et al}\onedot}
\makeatother

\usepackage[capitalize]{cleveref}
\crefname{section}{Sec.}{Secs.}
\Crefname{section}{Section}{Sections}
\Crefname{table}{Table}{Tables}
\crefname{table}{Tab.}{Tabs.}
\Crefname{equation}{Equation}{Equations}
\crefname{equation}{Eqn.}{Eqns.}

\usepackage{lipsum}
\usepackage{bm}
\usepackage{overpic}
\usepackage[american]{babel}
\begin{document}
% \listoffigures
\captionsetup[figure]{labelfont={bf},labelformat={default},labelsep=period,name={Fig.}}
%\bibliographystyle{IEEEtran}
%\bibliography{egbib}
\title{Learning Diverse Tone Styles for Image Retouching}

\author{Haolin Wang, Jiawei Zhang, Ming Liu, Xiaohe Wu, Wangmeng Zuo,~\IEEEmembership{Senior Member,~IEEE}
	% <-this % stops a space
	\thanks{Haolin Wang, Ming Liu, Xiaohe Wu, and Wangmeng Zuo are with the School of Computer Science and Technology, Harbin Institute of Technology, Harbin 150001, China (e-mail: \href{mailto:why_cs@outlook.com}{why\_cs@outlook.com}, \href{mailto:csmliu@outlook.com}{csmliu@outlook.com}, \href{mailto:xhwu.cpsl.hit@gmail.com}{xhwu.cpsl.hit@gmail.com}, \href{mailto:wmzuo@hit.edu.cn}{wmzuo@hit.edu.cn}).
		
		Jiawei Zhang is with the SenseTime Research, Shenzhen 518038, China (e-mail: \href{mailto:zhjw1988@gmail.com}{zhjw1988@gmail.com}).}}
%\thanks{This paper was produced by the IEEE Publication Technology Group. They are in Piscataway, NJ.}% <-this % stops a space
%\thanks{Manuscript received April 19, 2021; revised August 16, 2021.}}

\markboth{Journal of \LaTeX\ Class Files,~Vol.~14, No.~8, August~2021}%
{Shell \MakeLowercase{\textit{et al.}}: A Sample Article Using IEEEtran.cls for IEEE Journals}

\IEEEpubid{0000--0000/00\$00.00~\copyright~2021 IEEE}

\maketitle

\begin{abstract}
    Image retouching, aiming to regenerate the visually pleasing renditions of given images, is a subjective task where the users are with different aesthetic sensations.
    Most existing methods adopt a deterministic model to learn the retouching style from a specific expert, making it less flexible to meet diverse subjective preferences.
    Besides, the intrinsic diversity of an expert due to the targeted processing of different images is also deficiently described.
    To circumvent such issues, we propose to learn diverse image retouching with normalizing flow-based architectures.
    Unlike current flow-based methods which directly generate the output image, we argue that learning in a style domain could
    (i)~disentangle the retouching styles from the image content,
    (ii)~lead to a stable style presentation form, and
    (iii)~avoid the spatial disharmony effects.
    For obtaining meaningful image tone style representations, a joint-training pipeline is delicately designed, which is composed of a style encoder, a conditional RetouchNet, and the image tone style normalizing flow (TSFlow) module.
    In particular, the style encoder predicts the target style representation of an input image, which serves as the conditional information in the RetouchNet for retouching, while the TSFlow maps the style representation vector into a Gaussian distribution in the forward pass.
    After training, the TSFlow can generate diverse image tone style vectors by sampling from the Gaussian distribution.
    Extensive experiments on MIT-Adobe FiveK and PPR10K datasets show that our proposed method performs favorably against state-of-the-art methods and is effective in generating diverse results to satisfy different human aesthetic preferences.
    Source code and pre-trained models are publicly available at \url{https://github.com/SSRHeart/TSFlow}.
\end{abstract}

\vspace{-1em}

\section{Introduction}

\IEEEPARstart{W}{ith} the rapid development of photographic equipment like mobile phone cameras, taking photos has become a very common activity in recent years.
Therefore, the demand for image retouching to regenerate visually more pleasing renditions of the photos is also significantly increased.
A series of commercial software such as Adobe Photoshop and Lightroom are ready to use for professional photographers, where stunning images can be created with useful functions and appropriate parameters.
However, the operations require specialized skills and experience, and it is tedious and time-consuming to deal with a large collection of photos.
As a remedy, numerous image retouching methods~\cite{HDRNet,DPE,DUPE,DeepLPF,GleNet,CSRNet,RLC,StarEnhancer,QAGAN,ijcai2021cGAN} have been proposed to perform image enhancement automatically by learning from expert-retouched image pairs or exploring non-paired training data in an unsupervised manner.
But most of these methods learn the retouching style with a deterministic model and generate only a single retouching style, which greatly limits the practical applications.

\begin{figure}[t]
    \small
    \centering
    \renewcommand\tabcolsep{2pt}
    \begin{tabular}{ccc}
        \includegraphics[width=.3\linewidth]{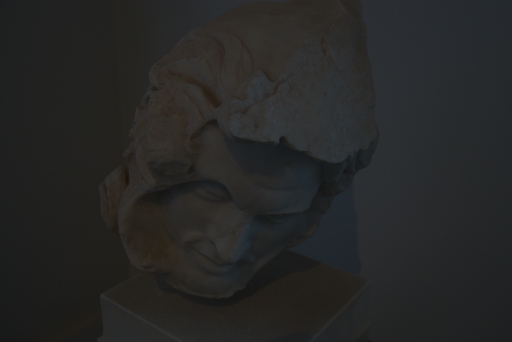} & \includegraphics[width=.3\linewidth]{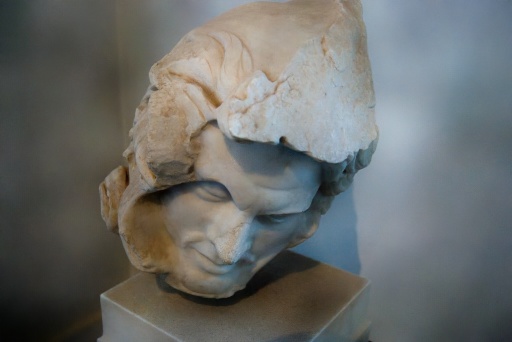} & \includegraphics[width=.3\linewidth]{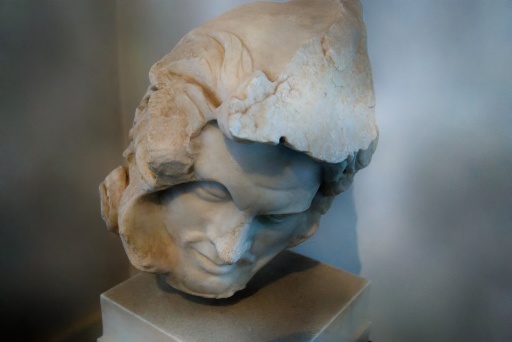} \\
        (a) Input & \multicolumn{2}{c}{(b) Diverse results (LLFlow)} \\
        % Input  & LLFlow (1) & LLFlow (2)
        \includegraphics[width=.3\linewidth]{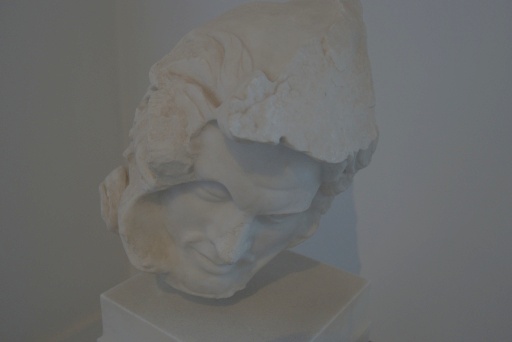} & \includegraphics[width=.3\linewidth]{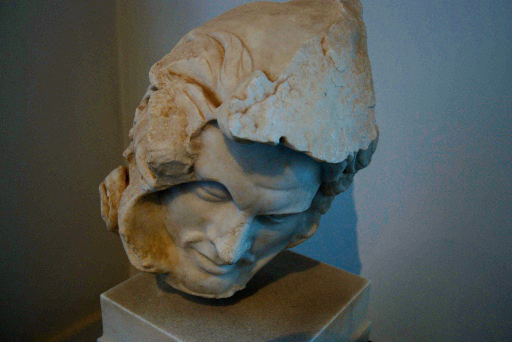} & \includegraphics[width=.3\linewidth]{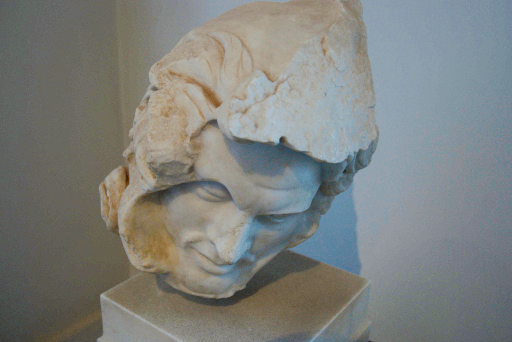} \\
        % Input (brightened) & Ours (1) & Ours (2)
        (c) Input (brightened) & \multicolumn{2}{c}{(d) Diverse results (Ours)} 
    \end{tabular}
    \caption{Diverse results generated by LLFlow~\cite{LLFlow} and our method. Spatial disharmony effects can be observed in the results of LLFlow, while our results are visually more pleasing. The image in (c) is the input with improved brightness for better observation.}
    \label{fig:ourvsLLFlow}
    \vspace{-1em}
\end{figure}

\IEEEpubidadjcol
In order to \textit{support additional retouching styles}, researchers have developed some fast adaptation strategies.
For example, CSRNet~\cite{CSRNet} fine-tunes only the lightweight conditional network with additional low-quality and expert-retouched pairs.
Sun~\etal~\cite{ijcai2021cGAN} leverage the low-quality input and an unpaired expert-retouched reference image for flexible inference.
Song~\etal~\cite{StarEnhancer} directly extract latent code from several images with a specific style via a style encoder.
Bianco~\etal~\cite{bianco2020personalized} extract the user preference profiles by fitting user-selected/retouched samples.
Since additional images with a specific style are required to generate that style, these methods are still far from convenient for ordinary users.
There are also some trials on \textit{providing extra unseen styles}.
CSRNet~\cite{CSRNet} shows the results of interpolating images generated with different styles and provides an interpolation parameter for manual control.
Thus, new styles can be created by fusing existing ones.
Kosugi~\etal~\cite{RLC} uses reinforcement learning to select the interpretable parameters in image retouching software (\eg, Adobe Photoshop), which builds a basis for further manual adjustment.
The former provides limited style space and does not get rid of the need for additional samples with extra styles, while the latter degenerates into complex and professional manual operations.

As one can see, even with the aforementioned efforts, current methods are less flexible to meet different subjective preferences, and there is still much room for generating diverse tone styles in image retouching tasks.
Besides, all previous methods regard the images retouched by an expert as a single style, which ignores the intrinsic diversity of the expert due to the targeted processing of different images. In this way, the model is actually learning an average style of the expert. On the contrary, we consider modeling the style space via a normalizing flow framework.
Theoretically, normalizing flow learns an invertible mapping between the tone style distribution and a simple Gaussian distribution.
In this way, the style of an expert is described by a distribution rather than a fixed point, and new tone styles can be obtained via randomly sampling in the Gaussian distribution.

In fact, normalizing flow has been successfully deployed for many tasks like image generation~\cite{GLOW}, image super-resolution~\cite{SRFlow}, image denoising~\cite{NoiseFlow}, \etc, which makes the models able to generate diverse results.
The randomness in these tasks is more like a local perturbation. It only influences detailed textures, which will not affect the principal content of the generated images.
However, for tasks that modulate the global feature of the images like image colorization~\cite{cFlow} and low-light image enhancement~\cite{LLFlow}, directly applying normalizing flow will cause severe spatial disharmony effects (see \cref{fig:ourvsLLFlow}).
Calculating the mean of different results can mitigate this effect but demolish the diversity~\cite{LLFlow} conversely.
To solve this problem, instead of directly mapping between images and the Gaussian distribution, we propose to learn the normalizing flow in a style domain, where the image tone style is represented via a latent vector.

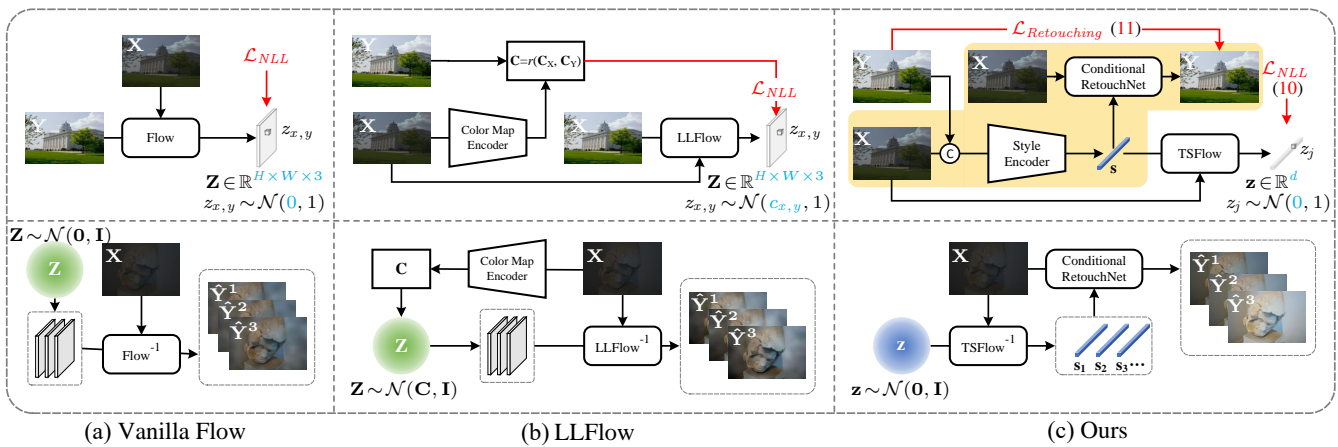
\begin{figure*}[t]
    \centering
    \scriptsize
    \begin{overpic}[width=.98\textwidth]{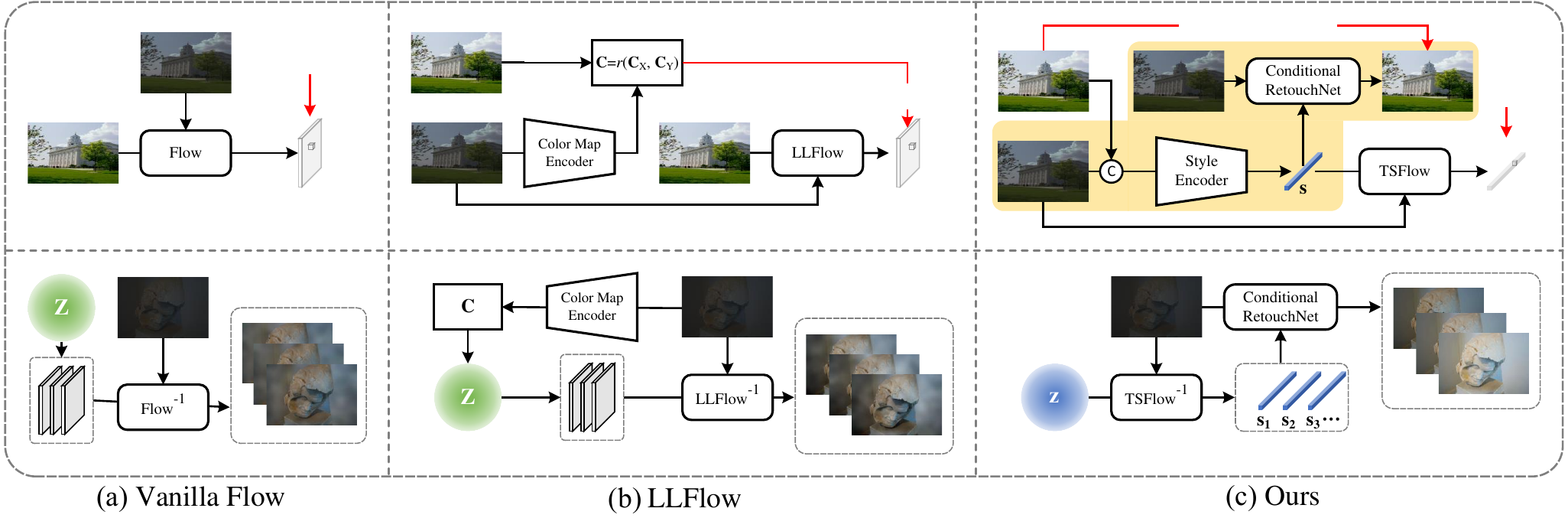}
        % TODO
        %		\put(0, 3){\color{red} TODO: $\mathbf{Y'}\rightarrow\mathbf{\hat{Y}}$,  $\mathbf{f}\rightarrow\mathbf{s}$, replace the images}
        
        % Vanilla Flow: Training
        \put(18, 29){\color{red}$\mathcal{L}_\mathit{NLL}$}
        \put(20.8, 23.5){$\mathit{z}_{\mathit{x},\mathit{y}}$}
        \put(15.2, 19.5){$\mathbf{Z}\!\in\!\mathbb{R}^{\color{cyan}\mathit{H}\times{W}\times3}$}
        \put(15, 18){$\mathit{z}_{\mathit{x},\mathit{y}}\!\sim\!\mathcal{N}({\color{cyan}0}, 1)$}
        \put(2, 23.7){\color{white}$\mathbf{Y}$}
        \put(9.3, 29.7){\color{white}$\mathbf{X}$}
        % Vanilla Flow: Inference
        \put(0.5, 15.5){$\mathbf{Z}\!\sim\!\mathcal{N}(\mathbf{0}, \mathbf{I})$}
        \put(15.5, 11.0){\color{white}$\mathbf{\hat{Y}^1}$}
        \put(16.2, 9.7){\color{white}$\mathbf{\hat{Y}^2}$}
        \put(17.0, 8.2){\color{white}$\mathbf{\hat{Y}^3}$}
        \put(7.7, 13.9){\color{white}$\mathbf{X}$}
        % LLFlow: Training
        % \put(35, 28.5){\tiny $\mathbf{C}\!=\!\mathit{r}(\mathbf{C}_\mathbf{X}, \mathbf{C}_\mathbf{Y})$}
        \put(56, 26.5){\color{red}$\mathcal{L}_\mathit{NLL}$}
        \put(58.8, 23.5){$\mathit{z}_{\mathit{x},\mathit{y}}$}
        \put(52.7, 19.5){$\mathbf{Z}\!\in\!\mathbb{R}^{\color{cyan}\mathit{H}\times{W}\times3}$}
        \put(51, 18){$\mathit{z}_{\mathit{x},\mathit{y}}\!\sim\!\mathcal{N}({\color{cyan}\mathit{c}_{\mathit{x},\mathit{y}}}, 1)$}
        \put(26.7, 23.7){\color{white}$\mathbf{X}$}
        \put(42.5, 23.7){\color{white}$\mathbf{X}$}
        \put(26.7, 29.7){\color{white}$\mathbf{Y}$}
        % LLFlow: Inference
        \put(26, 4){$\mathbf{Z}\!\sim\!\mathcal{N}(\mathbf{C}, \mathbf{I})$}
        
        \put(43.7, 13.9){\color{white}$\mathbf{X}$}
        \put(51.5, 10.3){\color{white}$\mathbf{\hat{Y}^1}$}
        \put(52.9, 9.0){\color{white}$\mathbf{\hat{Y}^2}$}
        \put(54.3, 7.7){\color{white}$\mathbf{\hat{Y}^3}$}
        % Ours: Training
        \put(75.5, 31){\color{red} $\mathcal{L}_\mathit{Retouching}$~{\color{black}(}\ref{eqn:loss_retouching}{\color{black})}}
        \put(94.2, 28){\color{red} $\mathcal{L}_\mathit{NLL}$}
        \put(95, 26.5){\color{red}{\color{black}(}\ref{eqn:loss_NLL}{\color{black})}}
        \put(97, 22){$\mathit{z}_{\mathit{j}}$}
        \put(93, 19.5){$\mathbf{z}\!\in\!\mathbb{R}^{\color{cyan}\mathit{d}}$}
        \put(91.5, 18){$\mathit{z}_{\mathit{j}}\!\sim\!\mathcal{N}({\color{cyan}0}, 1)$}
        \put(63.7, 22.7){\color{white}$\mathbf{X}$}
        \put(63.7, 28.4){\color{white}$\mathbf{Y}$}
        \put(72.5, 28.4){\color{white}$\mathbf{X}$}
        \put(88.2, 28.4){\color{white}$\hat{\mathbf{Y}}$}
        % Ours: Inference
        \put(63.5, 4){$\mathbf{z}\!\sim\!\mathcal{N}(\mathbf{0}, \mathbf{I})$}
        \put(71, 13.9){\color{white}$\mathbf{X}$}
        \put(89, 13.2){\color{white}$\mathbf{\hat{Y}^1}$}
        \put(90.3, 11.7){\color{white}$\mathbf{\hat{Y}^2}$}
        \put(91.6, 10.2){\color{white}$\mathbf{\hat{Y}^3}$}
    \end{overpic}
    \vspace{-0.3em}
    \caption{Illustration of the proposed image retouching framework. The pipeline of the vanilla normalizing flow-based baseline and LLFlow~\cite{LLFlow} are given for comparison. Details are depicted in \cref{sec:method_motivation}. The upper half is for the training phase, while the lower half is for inference. Even with a color map as the mean of the Gaussian distribution, LLFlow still faces the spatial disharmony problem. $\mathbf{X}$, $\mathbf{Y}$ and $\mathbf{\hat{Y}}$ are input, reference, and output images. $\mathit{z}_{\mathit{x}, \mathit{y}}$, $\mathit{c}_{\mathit{x}, \mathit{y}}$, $\mathit{z}_\mathit{j}$ denote a point of tensor $\mathbf{Z}$, tensor $\mathbf{C}$, and vector $\mathbf{z}$, respectively\protect\footnotemark. $\mathbf{s}$ is style vector. $\mathit{r}(\mathbf{A}, \mathbf{B})$ denotes random selection between $\mathbf{A}$ and $\mathbf{B}$. {\textcircled{{\textsf{c}}}} means concatenation operation. $\mathbf{I}$ denotes identity matrix. By learning the distribution of image tone style, our method eliminates the spatial disharmony effects. The yellow part shows a traditional deterministic image retouching model that degenerates from our training framework.}
    \label{fig:framework}
    \vspace{-0.7em}
\end{figure*}

In order to obtain meaningful image tone style representations, we have constructed a joint-training pipeline, which is composed of a style encoder, a conditional RetouchNet, and a tone style normalizing flow (TSFlow) module.
In particular, the style encoder predicts the target style vector $\mathbf{s}$ for an input image.
Then the RetouchNet takes $\mathbf{s}$ as conditional information for processing the input.
By constraining the output to approximate the expert-retouched reference, the style vector $\mathbf{s}$ should contain meaningful image tone style features, and now we can learn the distribution of the styles via the TSFlow module.
Interestingly, the style encoder and the conditional RetouchNet can form a traditional image retouching model (the yellow part in \cref{fig:framework}), where the image tone style vector of test images is still predicted by the deterministic style encoder.
On the contrary, during inference, we can obtain diverse image tone style vectors by sampling random noise vectors from the Gaussian distribution and mapping them to the style space by TSFlow. 
Therefore, the style encoder in our framework can be safely discarded after training.

It is worth noting that a key factor that influences the training process is the quality of the style representations extracted by the style encoder.
With our new framework, the style encoder is only used for training, thus we choose to take the expert-retouched reference image as another input of the style encoder.
In this way, the task of the style encoder becomes \textit{extracting} the style representation from the input-reference pair instead of \textit{predicting} solely from an input image, which greatly improves the style representation quality.
Note that such modification is also consistent with our proposal to describe the style of an expert with a distribution, and guarantees that the style the expert particularly designed for an image is fed into the TSFlow.
Besides, it enables the style encoder to support diverse styles.
A progressive style correction paradigm is also proposed to further refine the image tone style representation.

Extensive quantitative and qualitative experiments are conducted on two commonly used standard benchmarks (\ie, MIT-Adobe FiveK~\cite{fiveK} and PPR10K~\cite{PPR10K}), which show that our proposed method performs favorably against state-of-the-art methods, and can not only better describe the style distribution of a single/multiple expert(s), but achieve better flexibility on generating diverse unseen tone styles for image retouching.

To sum up, the contributions of this paper involve,
\begin{itemize}
    \item We propose a normalizing flow-based method to learn diverse tone styles for image retouching. By learning in the style space, the proposed method eliminates the spatial disharmony effects of existing normalizing flow-based methods.
    \item The image-specific style extraction and progressive style correction strategies are utilized for enhancing the tone style representation quality and modeling the intrinsic style diversity of an expert.
    \item Extensive experiments show that the proposed model can generate multiple retouching results to satisfy different subjective preferences and outperforms state-of-the-art methods quantitatively and qualitatively.
\end{itemize}

\section{Related Work}

In this section, we first review the current image retouching methods, which can be divided into two categories according to the number of styles supported, \ie, single-style and multi-style.
Then, the conditional normalizing flow-based methods relevant to this paper are also briefly introduced.

% ------------------------------------------------------------------------------
\subsection{Single-style Image Retouching}

Given a low-quality input, single-style image retouching methods produce a single deterministic result.
Early explorations include global transformation, histogram equalization, and Retinex-based methods.
Gamma correction and log function are widely used for global transformation.
Histogram equalization-based methods~\cite{kim1997contrast,stark2000adaptive,lee2011power,lee2013contrast} can modify the color histogram and expand the dynamic range of given images.
Retinex-based methods~\cite{fu2015probabilistic,stark2000adaptive,fu2016weighted,guo2016lime,yue2017contrast} chose to modify the illumination and preserve the reflectance of images.

Since the MIT-Adobe FiveK dataset was collected by Bychkovsky~\etal~\cite{fiveK}, many learning-based methods have been proposed to leverage the power of neural networks~\cite{HDRNet,DPE,li2020PiecewiseCurves,3DLUT,DeepLPF,QAGAN,PPR10K,LSDGT}.
Some methods~\cite{li2020PiecewiseCurves,GleNet} reformulated image retouching as a curve estimation task.
DeepLPF~\cite{DeepLPF} learns parameters of three types of local filters, which achieves local image retouching.
Considering the inference time and memory consumption in practical application, \cite{3DLUT,PPR10K} chose to learn image-adaptive three-dimensional lookup tables (3D LUTs) for efficient and satisfactory image retouching.
GANs are also utilized to achieve image retouching under the unpaired supervision~\cite{QAGAN,GleNet,DPE,UEGAN}, \eg, Ni~\etal~\cite{QAGAN} proposed a quality attention module to learn the semantic-related information between the low-quality and high-quality domains.
These methods can only generate a single-style result which is insufficient to satisfy different human aesthetic preferences.

% ------------------------------------------------------------------------------
\subsection{Multi-style Image Retouching}

As a remedy, some methods tried to generate multiple retouching results to cover diverse human aesthetic preferences~\cite{li2020PiecewiseCurves,RLC,PieNet,StarEnhancer}.
For example, He~\etal~\cite{CSRNet} proposed CSRNet to generate multiple expert styles by finetuning the conditional network for different ground truth references.
Sun~\etal~\cite{ijcai2021cGAN} proposed a lightweight conditional generative adversarial network to achieve one-to-many image enhancement with the guidance of various reference images.

To support unseen style preferences of new users, some methods~\cite{StarEnhancer,PieNet} proposed to estimate the different preference representations by providing a few images and then feeding different preference representations into a retouching network to generate results with additional unseen styles.
For example, Song~\etal~\cite{StarEnhancer} proposed StarEnhancer, where a tonal style classifier generates the center embedding of the specific style given some reference images.
Kim~\etal~\cite{PieNet} introduced metric learning to extract the preference vector from a few positive and negative images.
\cite{li2020PiecewiseCurves,RLC} provided a manner for generating extra results, which predicts explainable retouching parameters (\eg, color mapping curves, Adobe Photoshop parameters, \etc) that can be further adjusted by users.
However, all these methods require reference images to obtain unseen styles or need professional skills to modulate the results, which is still very inconvenient for ordinary users, so we propose to sample new styles with the normalizing flow-based framework.

% ------------------------------------------------------------------------------
\subsection{Conditional Normalizing Flow}
%\vspace{-3mm}
Normalizing flow~\cite{RealNVP,NICE,GLOW} is a powerful generative model with invertible structures, which maps a complex distribution into a simple one (\eg, Gaussian distribution) in the forward pass, and achieves diverse generations by sampling from the simple distribution in the reverse process.
With such characteristics, conditional normalizing flow-based models can be used to model the solution space of many ill-posed problems.
For example, SRFlow~\cite{SRFlow} takes the LR observation as the conditional information for generating diverse SR results.
Kim~\etal~\cite{NoiseSRFlow} further designed a noise conditional layer to improve the visual quality and diversity.
Liang~\etal~\cite{liang2021flow} proposed a normalizing flow-based kernel prior for kernel modeling, which can improve the performance of blind SR.
Abdelhamed~\etal~\cite{NoiseFlow} proposed a powerful noise modeling method that can produce realistic noise conditioned on the raw image and a series of camera parameters.
Abdal~\etal~\cite{StyleFlow} replaced the mapping network of StyleGAN2~\cite{StyleGAN2} with a conditional normalizing flow for image attribute editing.
Besides, normalizing flow is also utilized in more global tasks like image colorization~\cite{cFlow} and low-light enhancement~\cite{LLFlow}, however, spatial disharmony effect (see \cref{fig:ourvsLLFlow}) can be observed, leading to visually unpleasant results.

%In this work, we employ normalizing flow to learn
% ==============================================================================

\section{Proposed Method}

To generate spatially harmonious image retouching results with diverse tone styles, we propose a joint-training framework to model the image tone style distribution.
In \cref{sec:method_motivation}, we first explain the motivation in detail.
Then, we introduce the learning paradigm and the network structure in \cref{sec:method_framework,sec:method_structure}, respectively.
Finally, the learning objectives are formally defined in \cref{sec:method_training}.

\footnotetext{In this paper, unless specified, bold uppercase letters (\eg, $\mathbf{Z}$) represent tensors, and bold lowercase letters (\eg, $\mathbf{z}$) represent vectors.}

% ------------------------------------------------------------------------------
\subsection{Motivation}
\label{sec:method_motivation}

Denote by $\mathbf{X}$ and $\mathbf{Y}$ the input low-quality image and the expert-retouched reference image, existing image retouching methods aim to produce an output image $\mathbf{\hat{Y}}$ to approximate the tone style of the reference image $\mathbf{Y}$, \ie,
\begin{equation}
	\mathbf{\hat{Y}}=\mathit{G}(\mathbf{X};\theta_\mathit{G}),
	\label{eqn:traditional_method}
\end{equation}
where $\mathit{G}$ denotes the retouching model with parameters $\theta_\mathit{G}$.
In these methods, $\theta_\mathit{G}$ will be fixed once the training procedure is finished, which leads to a deterministic result.
However, the image retouching task is subjective, and the users have different aesthetic sensations, thus it is appealing to learn a model to generate various image retouching results.

For this purpose, normalizing flow~\cite{RealNVP,NICE,GLOW} is a reasonable choice, which achieves diverse generation with randomness by design.
But directly applying it for image retouching will cause severe spatial disharmony effects (\cref{fig:ourvsLLFlow}).
The reason is that the principal feature is provided by the conditional information (\eg, the low-resolution image in image super-resolution tasks), and the latent representation only encodes the local variations. In the image retouching task, the target style information is unavailable in the input image, so the tone style is dominated by the random noise sampled from the Gaussian distribution, \ie, $\mathbf{Z}\!\sim\!\mathcal{N}(\mathbf{0}, \mathbf{I})$.
The left column of \cref{fig:framework} shows a vanilla normalizing flow architecture applied for image retouching, where
\begin{equation}
	\mathbf{\hat{Y}}=\mathcal{F}^{-1}(\mathbf{Z}; \mathbf{X}, \theta_\mathcal{F}),
	\label{eqn:vanilla_flow}
\end{equation}
where $\mathcal{F}$ is the invertible structure of the normalizing flow.
It can be seen that $\mathbf{Z}\in\mathbb{R}^{\mathit{H}\times\mathit{W}\times3}$. Although all elements follow identical distributions, the inevitable spatial discrepancies make it difficult for the tone style to be consistent in different regions, and the varying latent representation shape further exacerbates the problem.
LLFlow~\cite{LLFlow} (the middle column of \cref{fig:framework}) provides more information by taking the predicted color map $\mathbf{C}$ as the mean of the Gaussian distribution, \ie, $\mathbf{Z}\!\sim\!\mathcal{N}(\mathbf{C}, \mathbf{I})$, yet the spatial disharmony problem still remains unsolved.

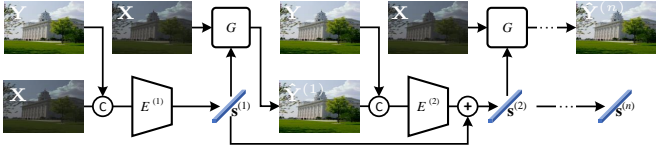
\begin{figure}[t]
    \centering
    % TODO
    \begin{overpic}[width=.98\linewidth]{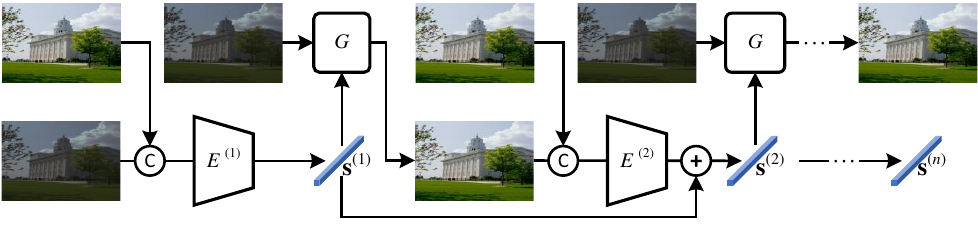}
        \put(0, 0){\color{red} }
        \put(1., 7.5){\color{white}\scriptsize $\mathbf{X}$}
        \put(1., 20.){\color{white}\scriptsize$\mathbf{Y}$}
        \put(17.5, 20.){\color{white}\scriptsize$\mathbf{X}$}
        
        \put(43., 7.5){\color{white}\scriptsize$\mathbf{\hat{Y}}^{(1)}$}
        \put(43., 20.){\color{white}\scriptsize$\mathbf{Y}$}
        \put(59.5, 20.){\color{white}\scriptsize$\mathbf{X}$}

        \put(88.5, 20.){\color{white}\scriptsize$\mathbf{\hat{Y}}^{(n)}$}    
    \end{overpic}  
    \caption{The progressive style correction paradigm. {\textcircled{{\textsf{+}}}} denotes the entry-wise addition operation. $\mathit{E}$ and $\mathit{G}$ denote the style encoder and RetouchNet, respecitively}
    \label{fig:progressive}
    \vspace{-1em}
\end{figure}

% ------------------------------------------------------------------------------
\subsection{Image Tone Style Representation Learning}
\label{sec:method_framework}

With the above analysis, we expect the normalizing flow to model only the \textit{global} tone style of an image, and propose to separate the image tone style from the content.
Specifically, the proposed model is composed of two steps for inference, \ie, first obtaining a style vector $\mathbf{s}$ with the tone style normalizing flow (TSFlow) module, then processing the input image via the RetouchNet $\mathit{G}$ with the guidance of $\mathbf{s}$.
The pipeline can be formulated as,
\begin{equation}
	\mathbf{\hat{Y}}=\mathit{G}(\mathbf{X}, \mathbf{s}; \theta_\mathit{G}),
	\label{eqn:RetouchNet}
\end{equation}
\begin{equation}
	\ \ \ \ \mathbf{s}=\mathcal{F}^{-1}(\mathbf{z}; \mathbf{X}, \theta_\mathcal{F}),
	\label{eqn:TSFlow_sampling}
\end{equation}
where $\mathbf{z}\!\sim\!\mathcal{N}(\mathbf{0}, \mathbf{I})$.

Such modification can benefit from several aspects.
To begin with, a separate style space can help the model to focus more on the image tone style rather than the content and to better capture the image tone style distribution.
Moreover, for the traditional two-dimensional normalizing flow framework, the spatial dimension of the latent representation will change with the input image size, and decomposing the style with the content will result in a more stable one-dimensional style representation form (\eg, a vector).
With the stable and disentangled style representation, the randomness is no longer able to cause local perturbations, and the spatial disharmony effects can be naturally eliminated.

For training such a framework, the key problem is to guarantee that $\mathbf{s}$ encodes meaningful image tone style created by the expert.
Some existing image retouching methods~\cite{StarEnhancer,CSRNet} follow similar configurations to represent the tone styles with a style vector, and achieve decent image retouching results.
Nevertheless, as illustrated in the yellow part of \cref{fig:framework}, to keep the consistency of the training and inference phase, these methods can only take the low-quality image as input of the style encoder, which is updated with the gradient of the loss calculated on the final output.
In other words, for a specific training pair ($\mathbf{X}$, $\mathbf{Y}$) from the training set ($\mathcal{X}^\mathit{tr}$, $\mathcal{Y}^\mathit{tr}$), the style vector $\mathbf{s}$ is predicted based on the overall style of the expert learned by the encoder $\mathit{E}$, rather than the specific style of the reference image $\mathbf{Y}$.
In this mode, the performance of learning the tone style distribution with normalizing flow drops, which is proved in \cref{ablation}.

\begin{figure}[t]
    \centering
    \begin{overpic}[width=.98\linewidth]{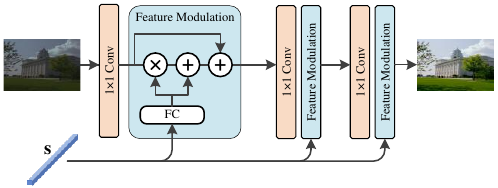}
        \put(0, 0){\color{red} }
        \put(1.5, 25.5){\color{white} $\mathbf{X}$}
       \put(84.5, 25.5){\color{white} $\hat{\mathbf{Y}}$}
    \end{overpic}		

    \vspace{-0.5em}
    \caption{Structure of the conditional RetouchNet. {\textcircled{{\small \textsf{$\times$}}}} denotes the entry-wise multiplication operation. More details are shown in \cref{sec:method_structure}}.
    \label{fig:RetouchNet}
    %	\vspace{-1.5em}
\end{figure}
\begin{figure}[t]
    \centering
    \begin{overpic}[width=.98\linewidth]{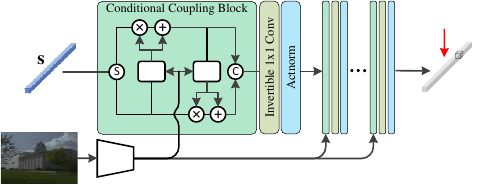}
        \put(86.2, 33){\color{red} $\mathcal{L}_\mathit{NLL}$}
        %		\put(95, 26.5){\color{red}{\color{black}(}\ref{eqn:loss_NLL}{\color{black})}}
        \put(93, 23){$\mathit{z}_{\mathit{j}}$}
        \put(84.5, 13.5){$\mathbf{z}\!\in\!\mathbb{R}^{\mathit{d}}$}
        \put(80.5, 9){$\mathit{z}_{\mathit{j}}\!\sim\!\mathit{N}(0, 1)$}
        % architecture
        \put(21.5, 4.2){$E^*$}
        \put(29.2, 22.5){$\varphi_1$}	
        \put(40.2, 22.5){$\varphi_2$}
        \put(0, 0){\color{red} }
        \put(1.5, 6.5){\color{white} $\mathbf{X}$}
    \end{overpic}		
    %	\vspace{-0.5em}
    \caption{Structure of the TSFlow.  {\textcircled{{\textsf{s}}}} means split operation and {\textcircled{{\textsf{c}}}} means concatenation on the channel dimension, respecitively. $E^*$ denotes the pretrained ResNet-18 network. More details are shown in \cref{sec:method_structure}}
    \label{fig:TSFlow}
    \vspace{-1em}
\end{figure}
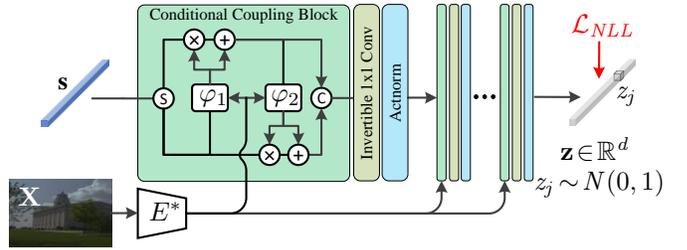

\vspace{0.5em}
\noindent\textbf{Image-specific Style Extraction}
Fortunately, as shown in \cref{eqn:TSFlow_sampling}, in our proposed framework, the tone style vector is sampled by the TSFlow during inference, so that the style encoder $\mathit{E}$ can be safely discarded after training.
Thus, we propose to \textit{take $\mathbf{Y}$ as another input of the style encoder}, \ie,
\begin{equation}
	\mathbf{s}=\mathit{E}(\mathbf{X}, \mathbf{Y}; \theta_\mathit{E}).
	\label{eqn:encoder}
\end{equation}
On the one hand, the expert will perform targeted processing for each image based on the understanding and experience, so that the reference image $\mathbf{Y}$ provides abundant information about the tone style, and the intrinsic diversity of an expert can be well described.
In such a way, the quality of the style representation is greatly enhanced.
On the other hand, since the style encoder is relieved from the burden of learning the style of a specific expert, we can utilize the reference retouching results from multiple experts to further enrich the tone style diversity.
%
%The proposed training framework is shown in \cref{fig:framework}, and the learning objectives will be described in \cref{sec:method_training}.
%In the following, we show the detailed design of each components.
%TODO: $\mathbf{Y'}\rightarrow\mathbf{\hat{Y}}$, $\mathbf{Y}^{(\mathit{n})}\rightarrow\mathbf{\hat{Y}}^{(\mathit{n})}$, replace the images
\begin{figure*}[t]
	%	\vspace{-4mm}
	\centering
	\begin{minipage}{0.15\linewidth}
		\includegraphics[width=\linewidth]{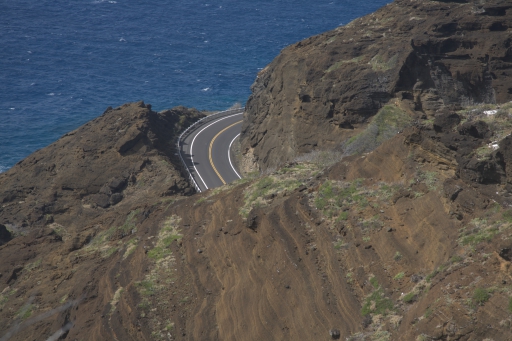}\\
		\vspace{-3mm}
		\includegraphics[width=\linewidth]{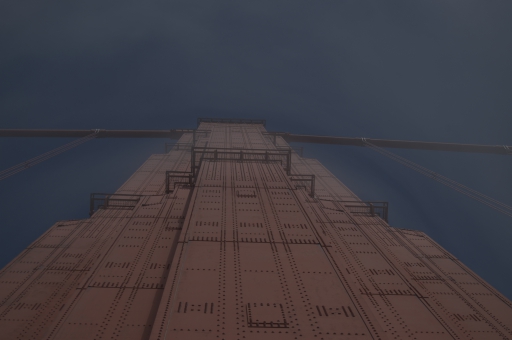}\\
		\vspace{-3mm}
		\includegraphics[width=\linewidth]{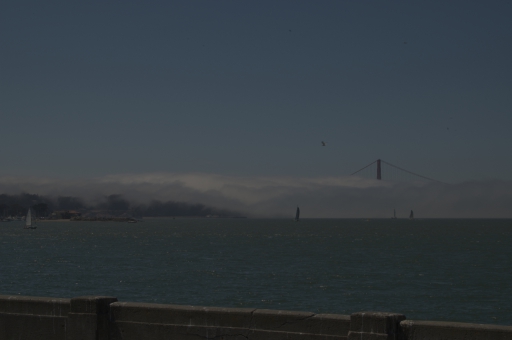}\\
		\vspace{-6mm}
		\caption*{Input}
	\end{minipage}
	\begin{minipage}{0.15\linewidth}
		\includegraphics[width=\linewidth]{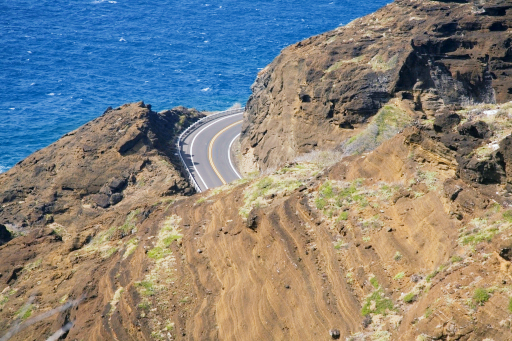}\\
		\vspace{-3mm}
		\includegraphics[width=\linewidth]{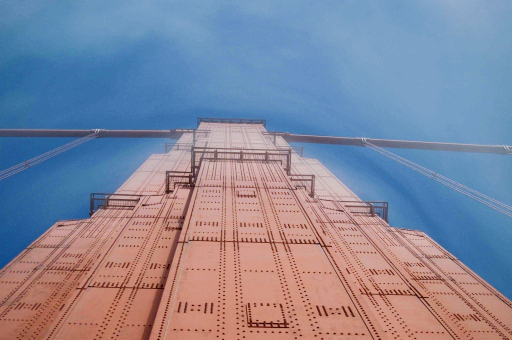}\\
		\vspace{-3mm}
		\includegraphics[width=\linewidth]{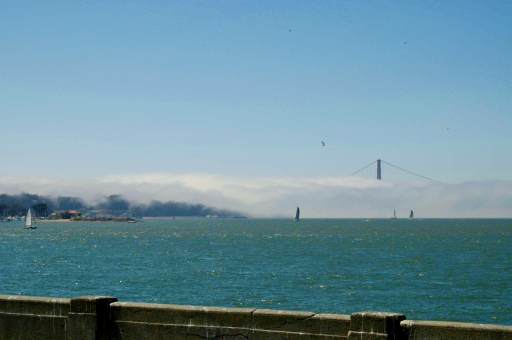}\\
		\vspace{-6mm}
		\caption*{CSRNet~\cite{CSRNet}}
	\end{minipage}
	\begin{minipage}{0.15\linewidth}
		\includegraphics[width=\linewidth]{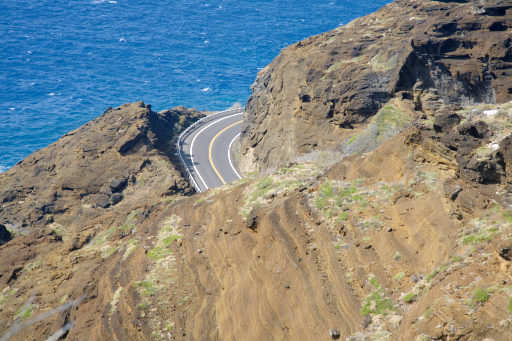}\\
		\vspace{-3mm}
		\includegraphics[width=\linewidth]{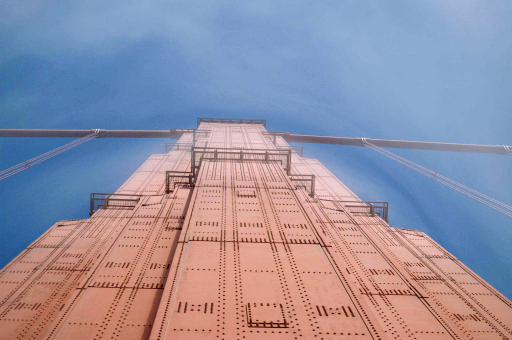}\\
		\vspace{-3mm}
		\includegraphics[width=\linewidth]{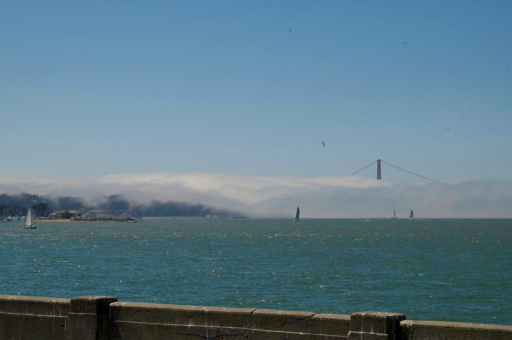}\\
		\vspace{-6mm}
		\caption*{3D LUT~\cite{3DLUT}}
	\end{minipage} 
	\begin{minipage}{0.15\linewidth}
		\includegraphics[width=\linewidth]{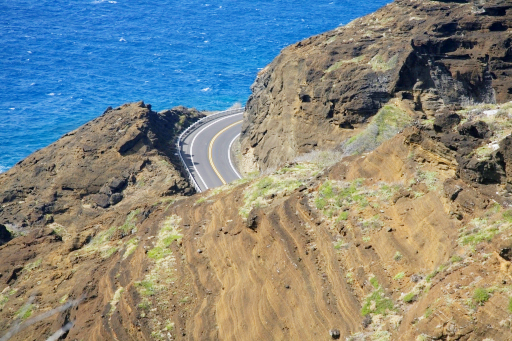}\\
		\vspace{-3mm}
		\includegraphics[width=\linewidth]{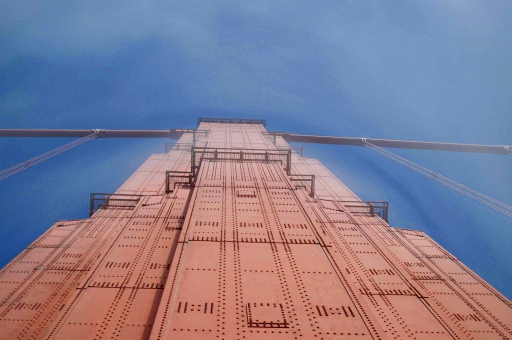}\\
		\vspace{-3mm}
		\includegraphics[width=\linewidth]{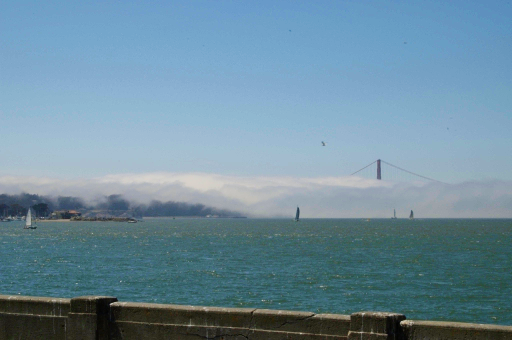}\\
		\vspace{-6mm}
		\caption*{StarEnhancer~\cite{StarEnhancer}}
	\end{minipage}
	\begin{minipage}{0.15\linewidth}
		\includegraphics[width=\linewidth]{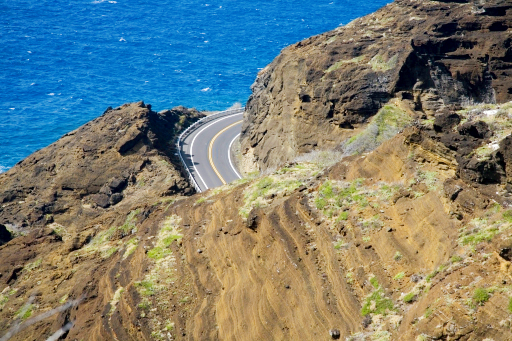}\\
		\vspace{-3mm}
		\includegraphics[width=\linewidth]{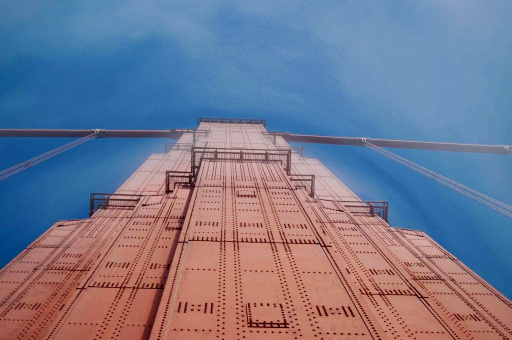}\\
		\vspace{-3mm}
		\includegraphics[width=\linewidth]{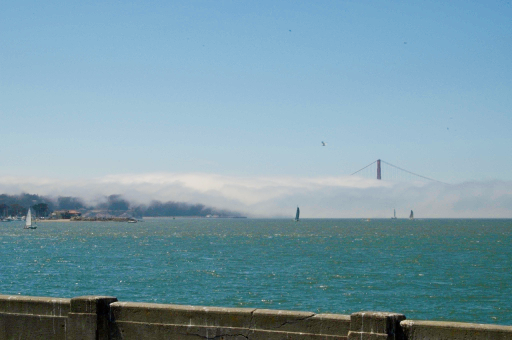}\\
		\vspace{-6mm}
		\caption*{Ours}
	\end{minipage}
	\begin{minipage}{0.15\linewidth}
		\includegraphics[width=\linewidth]{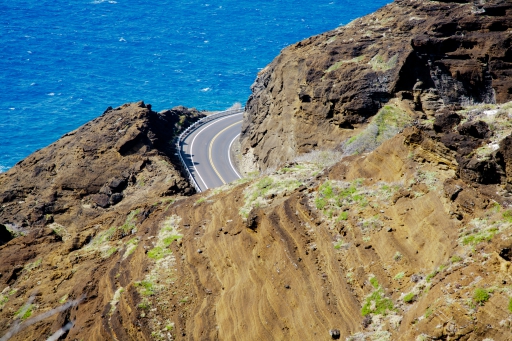}\\
		\vspace{-3mm}
		\includegraphics[width=\linewidth]{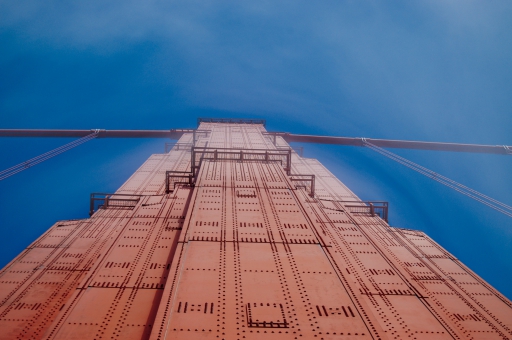}\\
		\vspace{-3mm}
		\includegraphics[width=\linewidth]{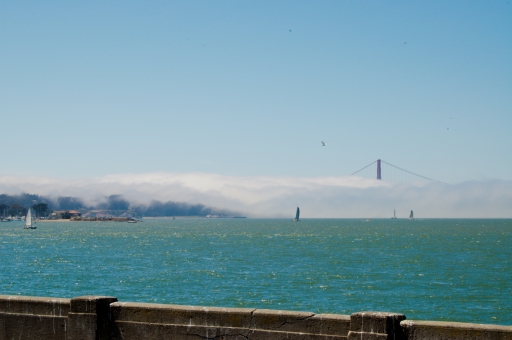}\\
		\vspace{-6mm}
		\caption*{Reference}
	\end{minipage}
	%	\vspace{24mm}
	\caption{Qualitative comparison on MIT-Adobe FiveK dataset. Results generated by our model are more similar with reference on image tone style. }
	\label{fig:SOTA fiveK}
	%	\vspace{-4mm}
\end{figure*}
\begin{figure*}[t]
	\centering
	\begin{minipage}{0.18\linewidth}
		\includegraphics[width=\linewidth]{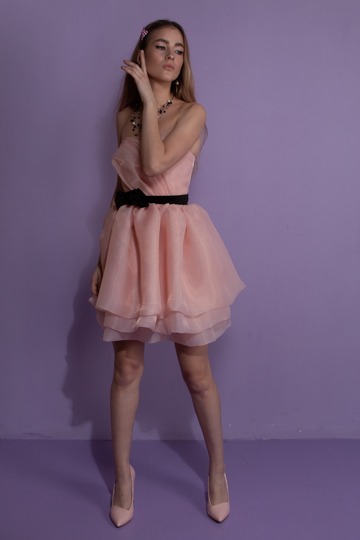}\\
		\vspace{-3mm}
		\includegraphics[width=\linewidth]{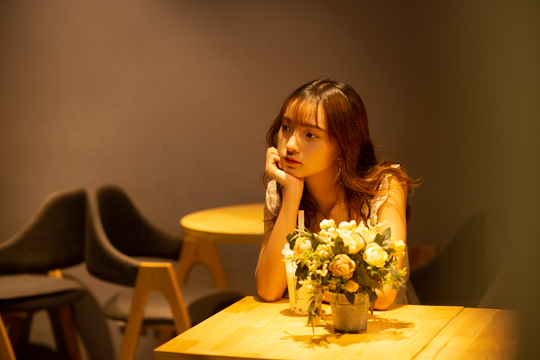}\\
		\vspace{-3mm}
		\includegraphics[width=\linewidth]{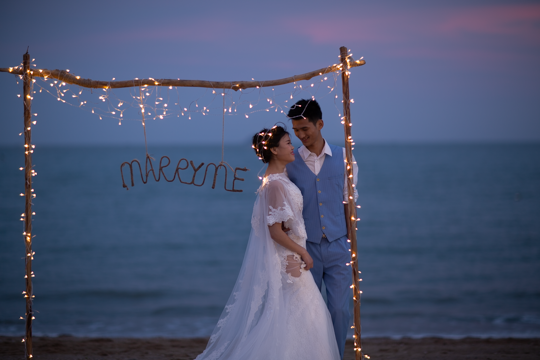}
		\caption*{Input}
	\end{minipage}
	\begin{minipage}{0.18\linewidth}
		\includegraphics[width=\linewidth]{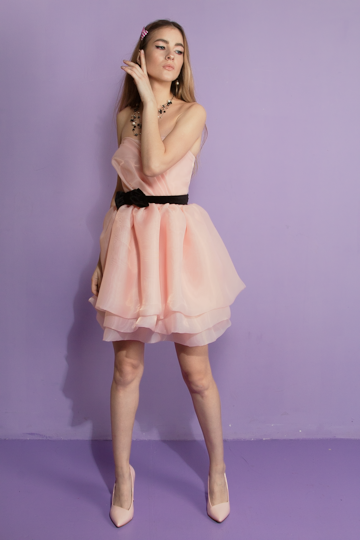}\\
		\vspace{-3mm}
		\includegraphics[width=\linewidth]{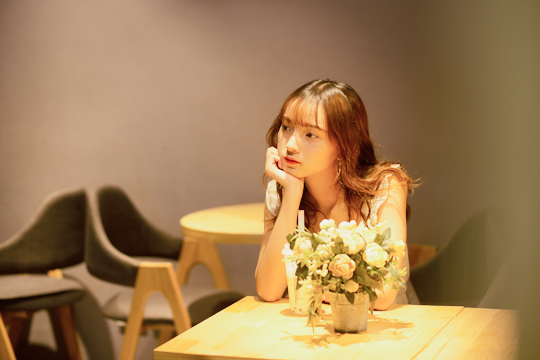}\\
		\vspace{-3mm}
				\includegraphics[width=\linewidth]{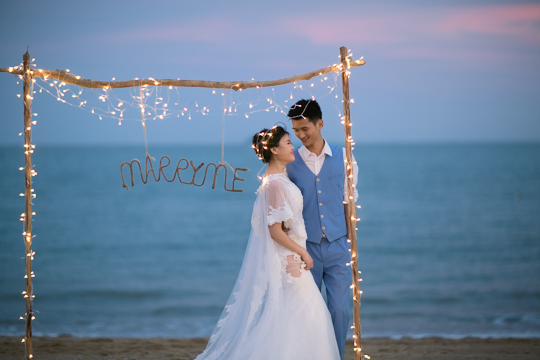}
		\caption*{CSRNet~\cite{CSRNet}}
	\end{minipage}
	\begin{minipage}{0.18\linewidth}
		\includegraphics[width=\linewidth]{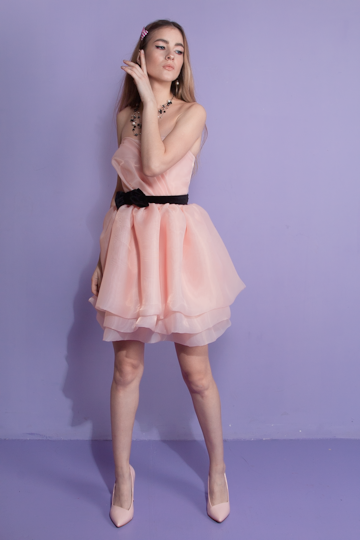}\\
		\vspace{-3mm}
		\includegraphics[width=\linewidth]{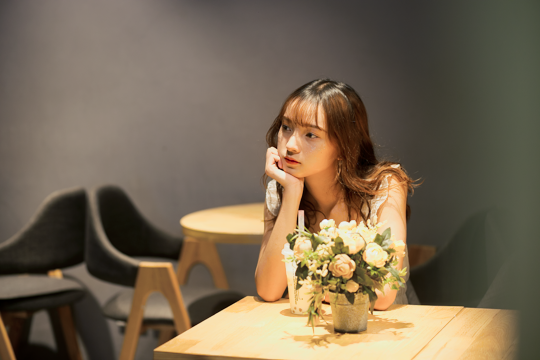}\\
		\vspace{-3mm}
				\includegraphics[width=\linewidth]{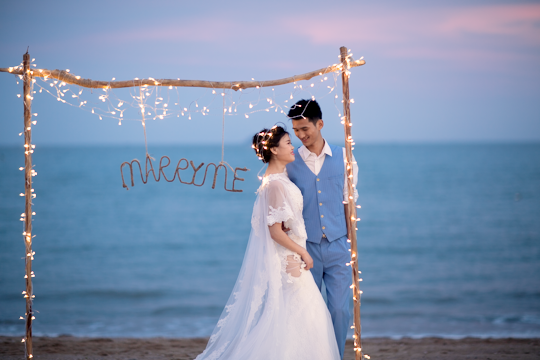}
		\caption*{3D LUT~\cite{3DLUT}}
	\end{minipage}
	\begin{minipage}{0.18\linewidth}
		\includegraphics[width=\linewidth]{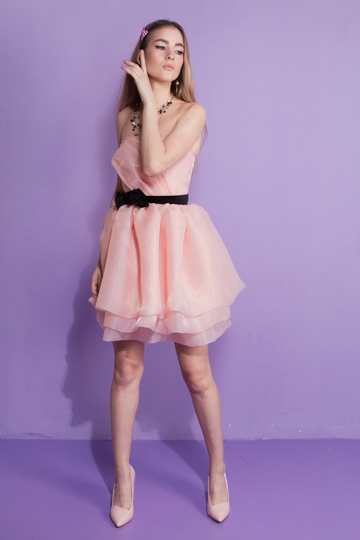}\\
		\vspace{-3mm}
		\includegraphics[width=\linewidth]{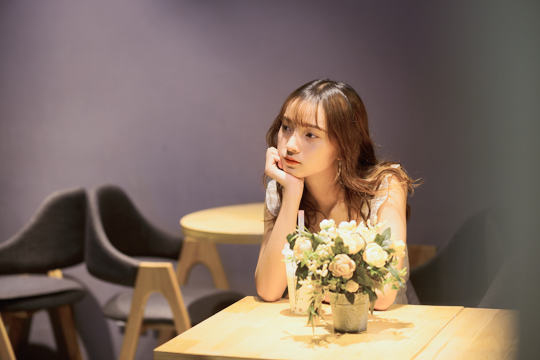}\\
		\vspace{-3mm}
				\includegraphics[width=\linewidth]{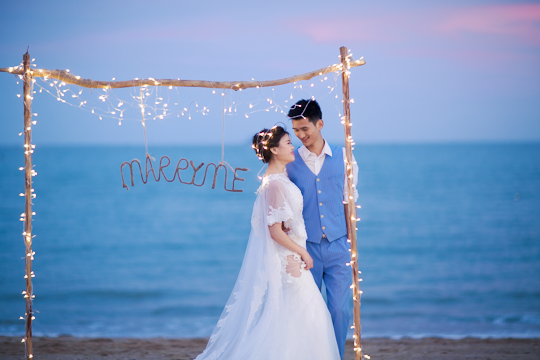}
		\caption*{Ours}
	\end{minipage}
	\begin{minipage}{0.18\linewidth}
		\includegraphics[width=\linewidth]{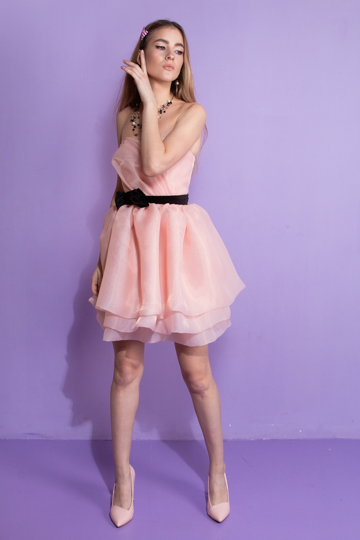}\\
		\vspace{-3mm}
		\includegraphics[width=\linewidth]{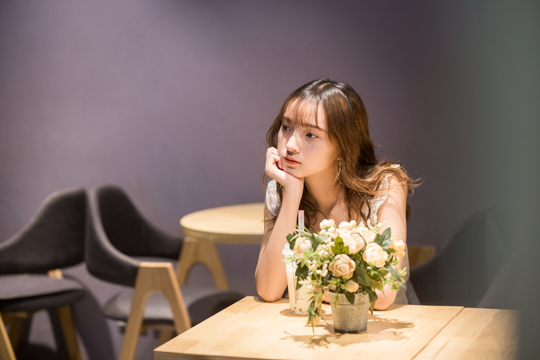}\\
		\vspace{-3mm}
				\includegraphics[width=\linewidth]{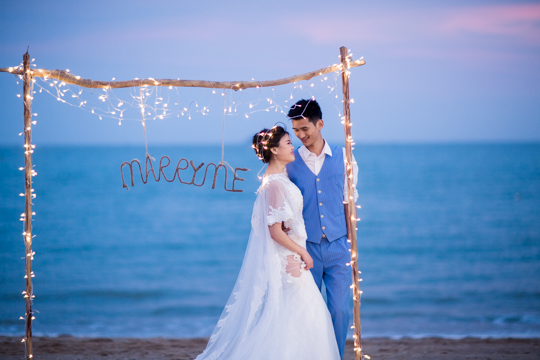}
		\caption*{Reference}
	\end{minipage}
	%	\vspace{24mm}
	\caption{Qualitative comparison on PPR10K dataset. Results generated by our model are more similar with reference on image tone style}
	\label{fig:sota PPR10K}
	%	\vspace{-4mm}
\end{figure*}

\vspace{0.5em}
\noindent\textbf{Progressive Style Correction}
To further improve the image tone style representation quality, we introduce a $\mathit{n}$-step progressive style correction paradigm to refine the image tone style vector.
As shown in \cref{fig:progressive}, the pipeline for each step is similar to the previous single-step one, except that the input at each step ($\mathit{t}$) is the retouching results of the previous step ($\mathit{t}-1$), \ie,
\begin{align}
	\mathbf{s}^{(\mathit{t})} &= \mathbf{s}^{(\mathit{t}-1)}+ \mathit{E}(\mathbf{\hat{Y}}^{(\mathit{t}-1)},\mathbf{Y};\theta_{\mathit{E}}^{(\mathit{t})}),\\
	\mathbf{\hat{Y}}^{(\mathit{t})} &= \mathit{G}(\mathbf{X},\mathbf{s}^{(\mathit{t})}; \theta_\mathit{G}),
\end{align}
where $\mathbf{s}^{(0)}=\mathbf{0}$, $\mathbf{\hat{Y}}^{(0)}=\mathbf{X}$, $\mathit{t}=1,\dots,n$.
%And for simplicity, we define $\mathbf{\hat{Y}}=\mathbf{\hat{Y}}^{(\mathit{n})}$ and $\mathbf{s}=\mathbf{s}^{(\mathit{n})}$.
In this paper, we find that $\mathit{n}=3$ is enough for the image tone style to achieve decent quality.

\begin{figure*}[!h]
    \centering
    \begin{minipage}{0.18\linewidth}
        \rotatebox{90}{\bfseries{Model trained on MIT-Adobe FiveK dataset}}
    \end{minipage}
    \hspace{-30mm}
    \begin{minipage}{0.18\linewidth}
        {\includegraphics[width=\linewidth]{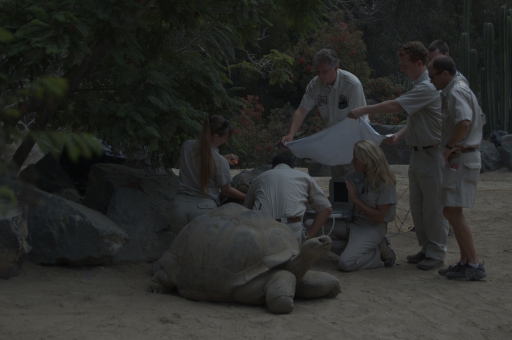}\\
            \vspace{-3mm}
            \includegraphics[width=\linewidth]{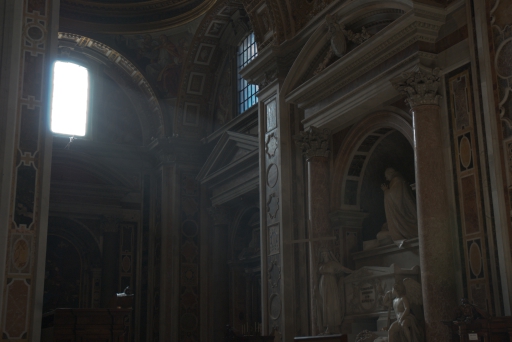}\\
            \vspace{-3mm}
            \includegraphics[width=\linewidth]{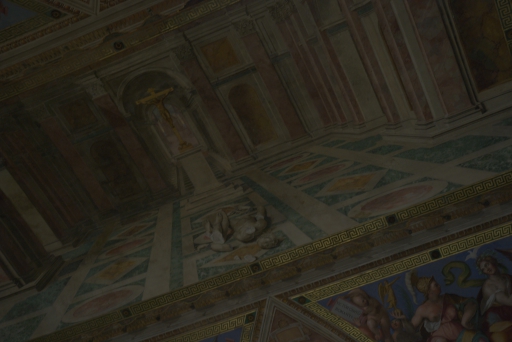}\\
            \vspace{-3mm}
            \includegraphics[width=\linewidth]{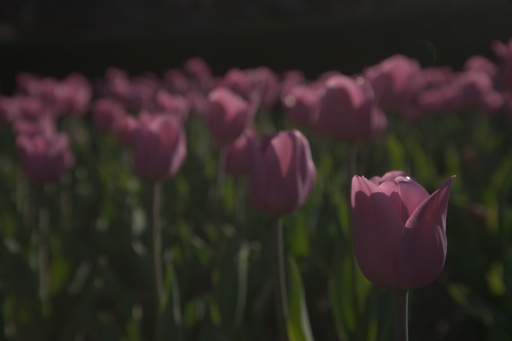}}
        \caption*{Input}
    \end{minipage}
    %	\vspace{-1mm}
    \begin{minipage}{0.18\linewidth}
        {\includegraphics[width=\linewidth]{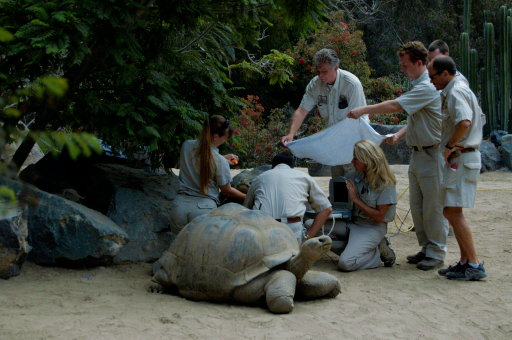}\\
            \vspace{-3mm}
            \includegraphics[width=\linewidth]{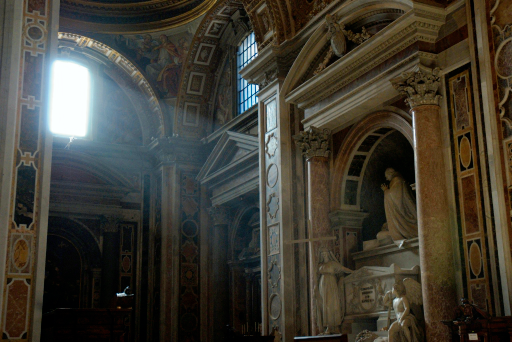}\\
            \vspace{-3mm}
            \includegraphics[width=\linewidth]{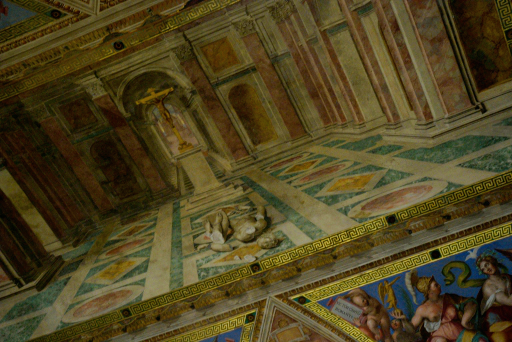}\\
            \vspace{-3mm}
            \includegraphics[width=\linewidth]{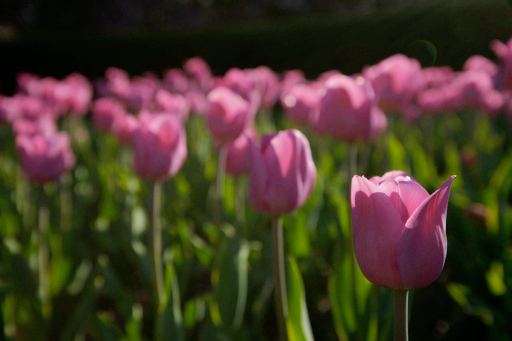}}
        \caption*{Retouching style 1}
    \end{minipage}
    \begin{minipage}{0.18\linewidth}
        {\includegraphics[width=\linewidth]{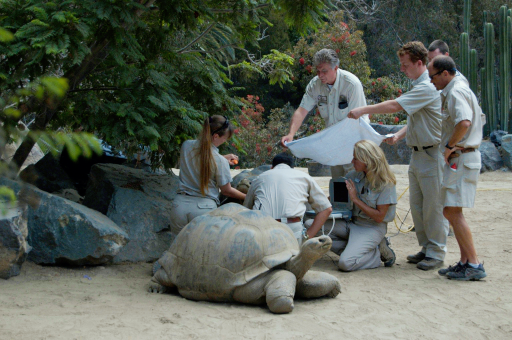}\\
            \vspace{-3mm}
            \includegraphics[width=\linewidth]{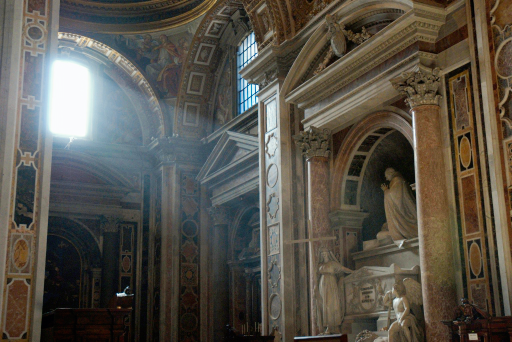}\\
            \vspace{-3mm}
            \includegraphics[width=\linewidth]{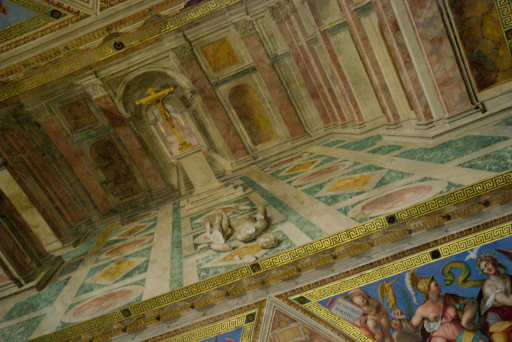}\\
            \vspace{-3mm}
            \includegraphics[width=\linewidth]{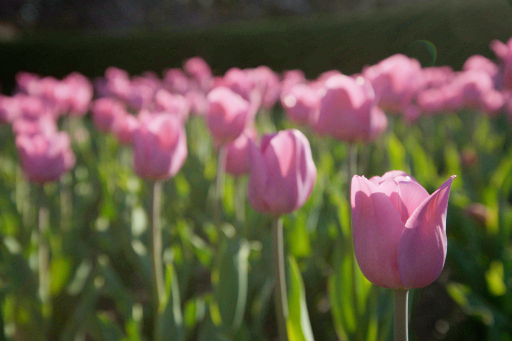}}
        \caption*{Retouching style 2}
    \end{minipage}
    \begin{minipage}{0.18\linewidth}
        {\includegraphics[width=\linewidth]{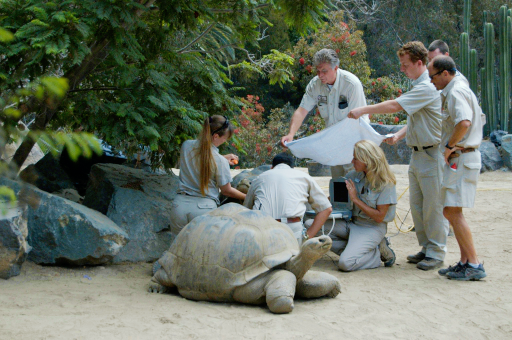}\\
            \vspace{-3mm}
            \includegraphics[width=\linewidth]{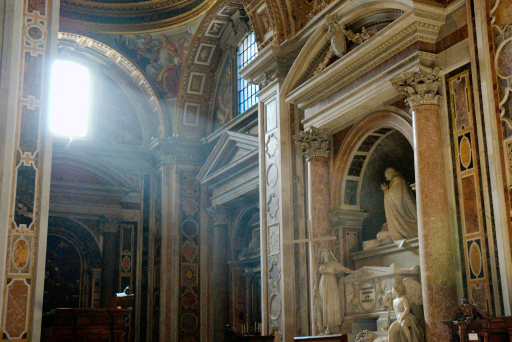}\\
            \vspace{-3mm}
            \includegraphics[width=\linewidth]{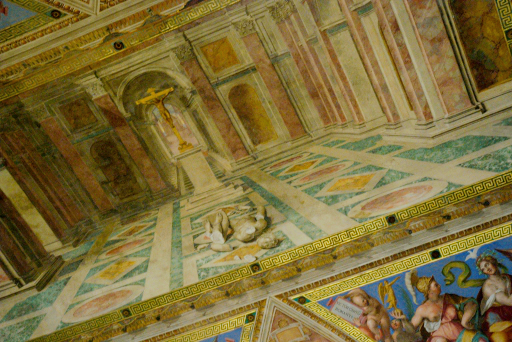}\\
            \vspace{-3mm}
            \includegraphics[width=\linewidth]{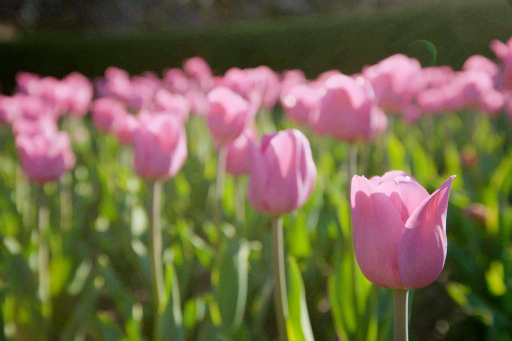}}
        \caption*{Retouching style 3}
    \end{minipage}
    \begin{minipage}{0.18\linewidth}
        {\includegraphics[width=\linewidth]{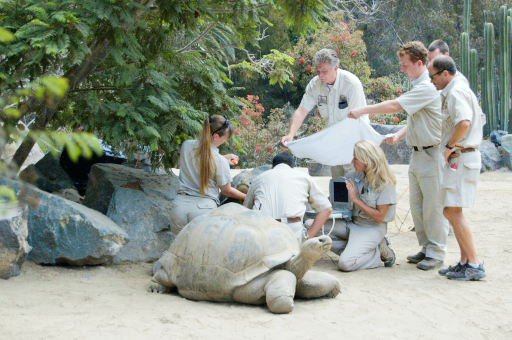}\\
            \vspace{-3mm}
            \includegraphics[width=\linewidth]{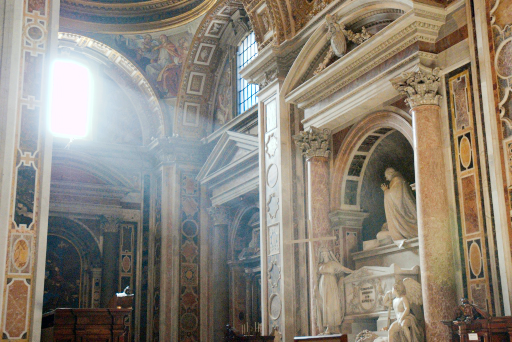}\\
            \vspace{-3mm}
            \includegraphics[width=\linewidth]{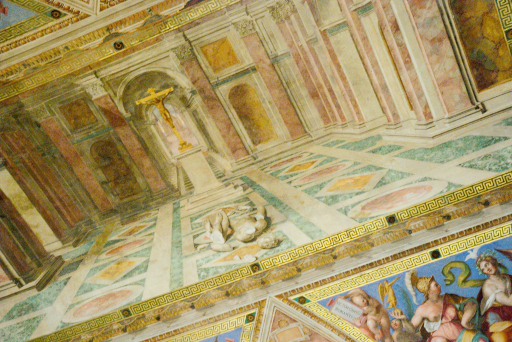}\\
            \vspace{-3mm}
            \includegraphics[width=\linewidth]{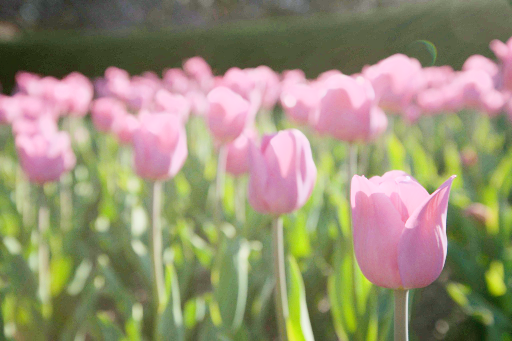}}
        \caption*{Retouching style 4}
    \end{minipage}
    
    %	\cdashline[0.5pt/5pt]
    
    \vspace{4mm}
    \begin{minipage}{0.18\linewidth}
        \rotatebox{90}{\bfseries{Model trained on PPR10K dataset}}
    \end{minipage}
    \hspace{-30mm}
    \begin{minipage}{0.18\linewidth}
        \includegraphics[width=\linewidth]{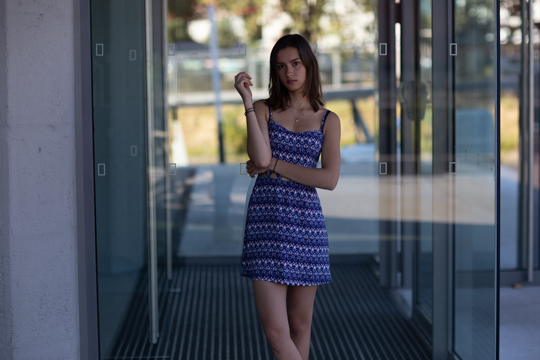}\\
        \vspace{-3mm}
        \includegraphics[width=\linewidth]{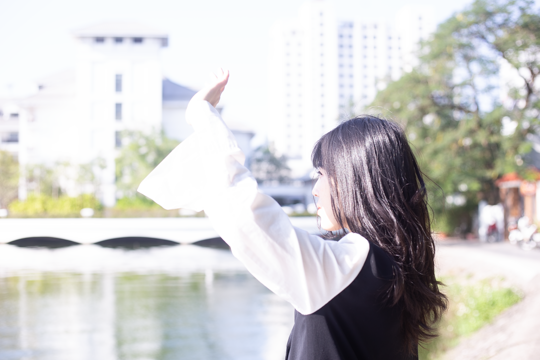}\\
        \vspace{-3mm}
        \includegraphics[width=\linewidth]{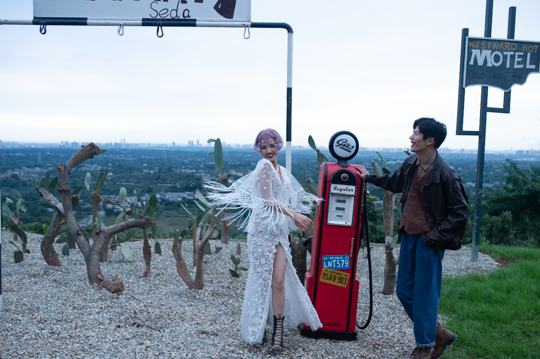}\\
        \vspace{-3mm}
        \includegraphics[width=\linewidth]{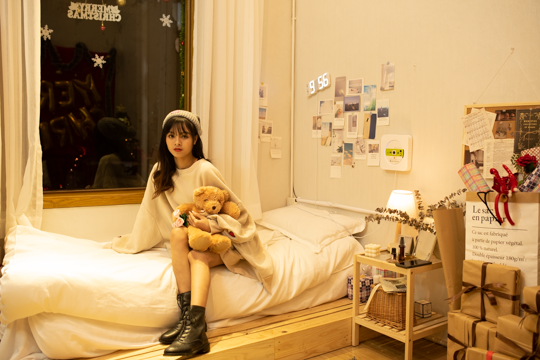}	
        \caption*{Input}
    \end{minipage}
    \begin{minipage}{0.18\linewidth}
        \includegraphics[width=\linewidth]{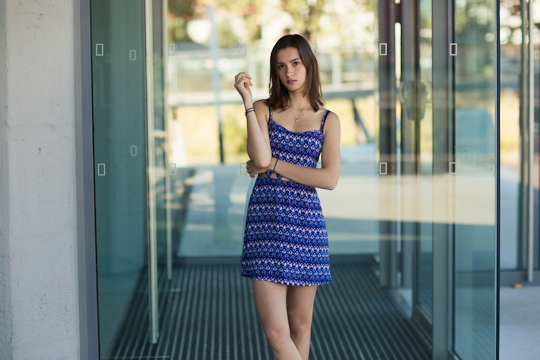}\\
        \vspace{-3mm}
        \includegraphics[width=\linewidth]{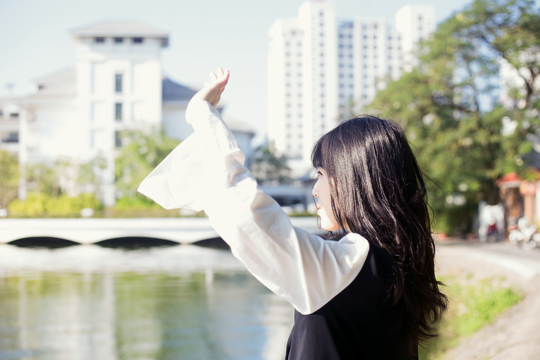}\\
        \vspace{-3mm}
        \includegraphics[width=\linewidth]{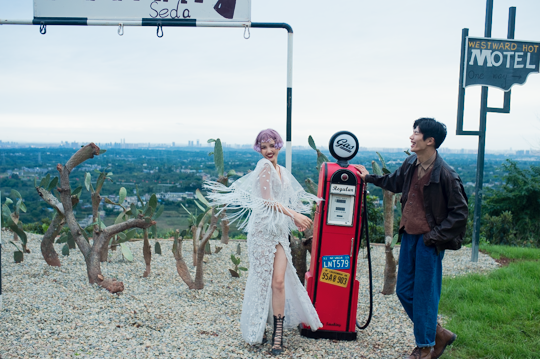}\\
        \vspace{-3mm}
        \includegraphics[width=\linewidth]{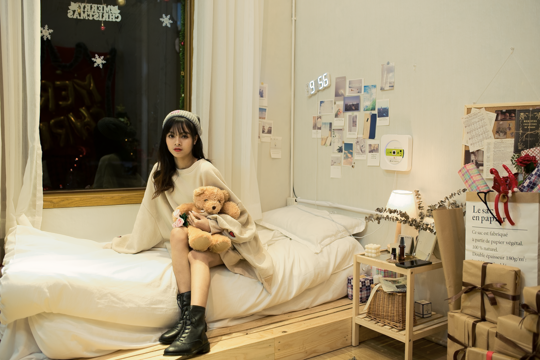}	  	
        \caption*{Retouching style 1}
    \end{minipage}
    \begin{minipage}{0.18\linewidth}	  	
        \includegraphics[width=\linewidth]{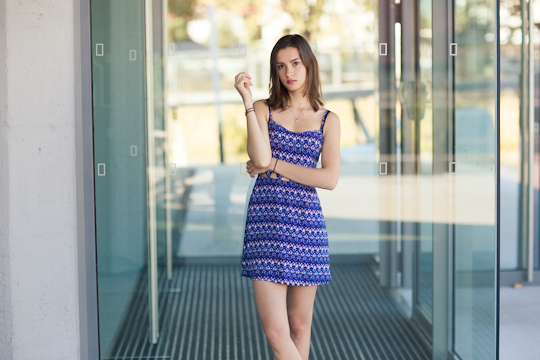}\\
        \vspace{-3mm}
        \includegraphics[width=\linewidth]{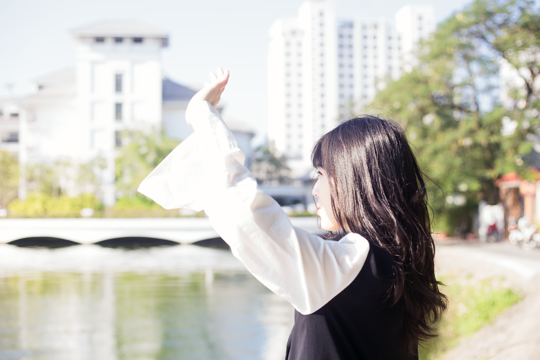}\\
        \vspace{-3mm}
        \includegraphics[width=\linewidth]{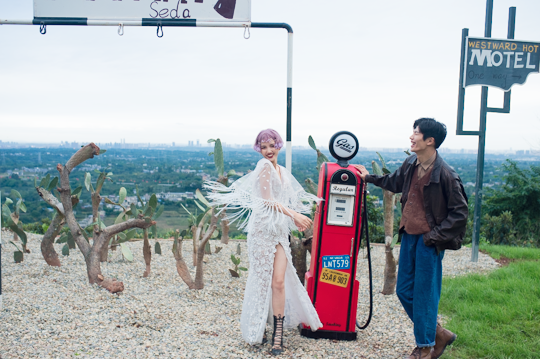}\\
        \vspace{-3mm}
        \includegraphics[width=\linewidth]{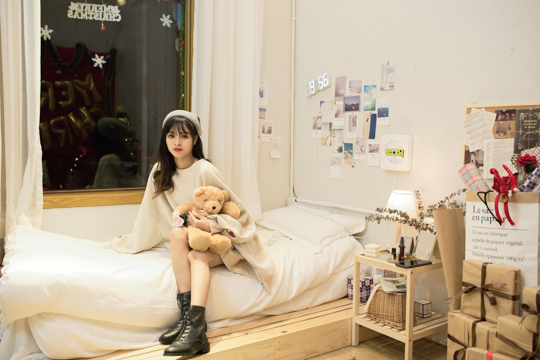}	  
        \caption*{Retouching style 2}
    \end{minipage}
    \begin{minipage}{0.18\linewidth}
        \includegraphics[width=\linewidth]{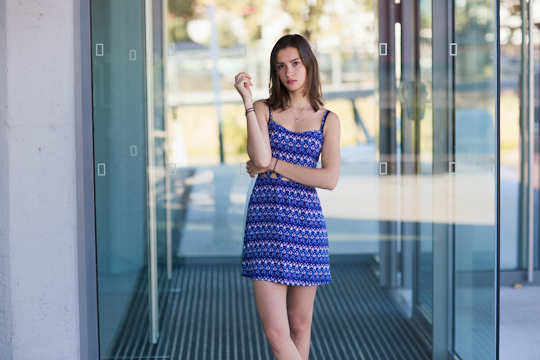}\\
        \vspace{-3mm}
        \includegraphics[width=\linewidth]{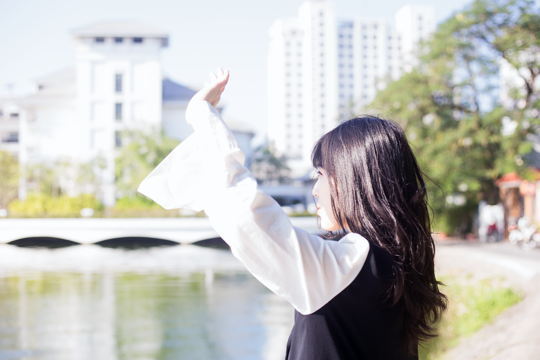}\\
        \vspace{-3mm}
        \includegraphics[width=\linewidth]{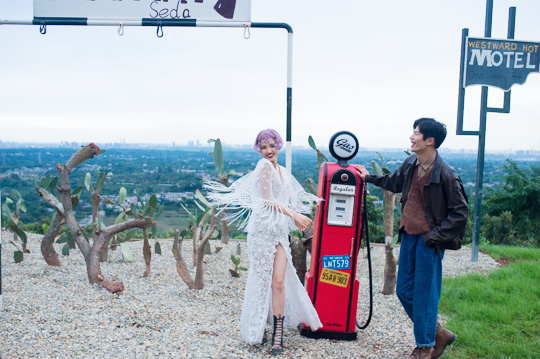}\\
        \vspace{-3mm}
        \includegraphics[width=\linewidth]{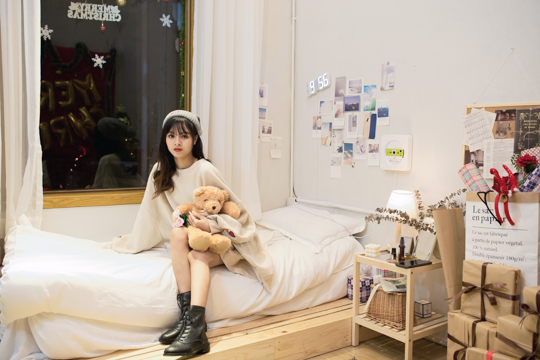}	  
        \caption*{Retouching style 3}
    \end{minipage}
    \begin{minipage}{0.18\linewidth}
        \includegraphics[width=\linewidth]{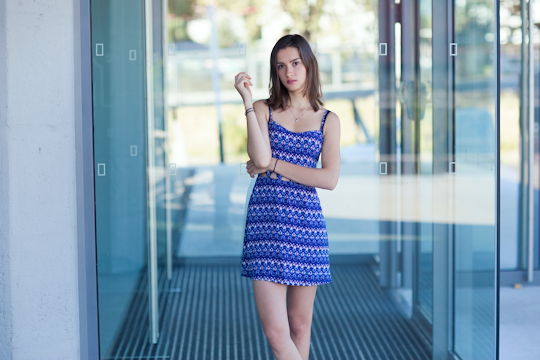}\\
        \vspace{-3mm}
        \includegraphics[width=\linewidth]{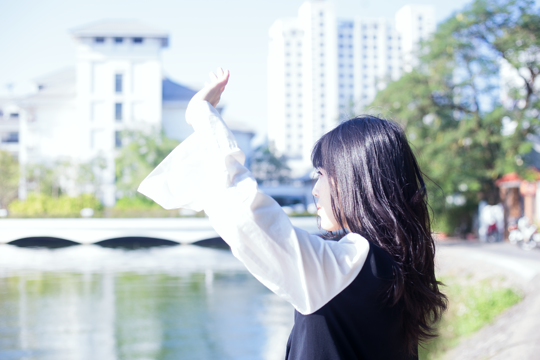}\\
        \vspace{-3mm}
        \includegraphics[width=\linewidth]{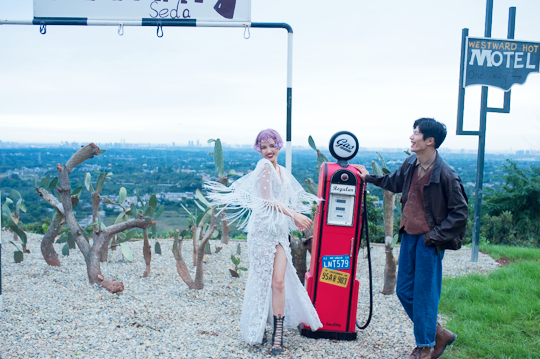}\\
        \vspace{-3mm}
        \includegraphics[width=\linewidth]{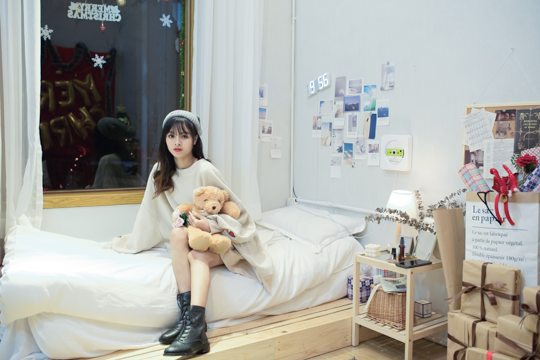}	  
        \caption*{Retouching style 4}
    \end{minipage}
    \caption{Diverse retouching results of models trained on MIT-Adobe FiveK dataset and PPR10K dataset, respectively.}
    \label{fig:diverseretouching}
\end{figure*}

% ------------------------------------------------------------------------------
\subsection{Network Structure}
\label{sec:method_structure}

As shown in \cref{eqn:RetouchNet,eqn:TSFlow_sampling,eqn:encoder}, the proposed framework has three components, \ie, a conditional RetouchNet $\mathit{G}$, a style encoder $\mathit{E}$, and the TSFlow $\mathcal{F}$.
The training and inference pipelines have been shown in \cref{fig:framework}.
In the following, we introduce the structure of these components in detail.

\vspace{0.5em}
\noindent\textbf{Conditional RetouchNet}
The conditional RetouchNet $\mathit{G}$ aims to reconstruct the expert-retouched result $\mathbf{\hat{Y}}$ from the low-quality input $\mathbf{X}$ with the guidance of a style representation $\mathbf{s}$.
As shown in \cref{fig:RetouchNet}, we design the RetouchNet as a lightweight model which contains three $1\!\times\!1$ convolution layers, each followed by a feature modulation module which can be formulated as,
\begin{equation}
    \mathbf{f}_\mathit{i}'=\mathbf{f}_\mathit{i}+(\alpha_\mathit{i}\mathbf{f}_\mathit{i}+\beta_\mathit{i}),
\end{equation}
where $\mathit{i}$ indicates the $\mathit{i}$-th channel of the feature map $\mathbf{f}$ or the $\mathit{i}$-th element of the scale / shift parameters $\bm{\alpha}$ and $\bm{\beta}$. Here $\bm{\alpha}$ and $\bm{\beta}$ are predicted from $\mathbf{s}$ by two FC layers to modulate the intermediate feature $\mathbf{f}$.

\vspace{0.5em}
\noindent\textbf{Style Encoder}
Since the style encoder $\mathit{E}$ can be discarded after training, and there is no need to consider the over-fitting problem of $\mathit{E}$, we can deploy a model with a large capacity for better capturing the style representations.
In practice, a network consisting of residual blocks and pooling layers~\cite{StarEnhancer} is applied for each step of the progressive style correction paradigm, where the input channel is 6 for both the input image $\mathbf{X}$ (or the retouching result of the prior step $\mathbf{\hat{Y}}^{(\mathit{t}-1)}$) and the expert-retouched reference image $\mathbf{Y}$.

\vspace{0.5em}
\noindent\textbf{TSFlow\label{sec:method_TSFlow}}
%
%\color{red} ResNet-18~\cite{ResNet} based architecture 
The TSFlow, denoted by $\mathcal{F}$, aims to learn the bijective mapping between the image tone style distribution and a simple Gaussian distribution conditioned on the given low-quality image $\mathbf{X}$.
Following previous conditional normalizing flow-based methods~\cite{cFlow,LLFlow,SRFlow}, our TSFlow is comprised of a sequence of flow steps, which contains a conditional coupling block~\cite{cFlow}, an invertible $1\times1$ convolution, and an actnorm layer.
Both the actnorm layer and the invertible $1\times1$ convolution layer follow the design of Glow~\cite{GLOW}.
As for the conditional coupling block, we adopt a pre-trained ResNet-18~\cite{ResNet} model to extract the condition information from the input image $\mathbf{X}$. Then the internal networks $\varphi_1$ and $\varphi_2$ take both the condition information and a half of the style code $\mathbf{s}$ as input and then achieve powerful transformation on the other half of $\mathbf{s}$ (see \cref{fig:TSFlow}). Note that both $\varphi_1$ and $\varphi_2$ are comprised of fully $1\times1$ convolution layers.

\begin{table}[t]
	\small
	\renewcommand\arraystretch{1.2}%.25}
	\caption{Quantitative comparison on {MIT-Adobe FiveK dataset}. The best results are shown in \textbf{bold}.}
	\label{tab:quantitative results}
	
	\begin{center}
		\begin{tabular}{cccc}
			%			\toprule
			\toprule
			Method &  $PSNR  \uparrow$ & $SSIM  \uparrow$   & $LPIPS  \downarrow$  \\
			\midrule
			HDRNet~\cite{HDRNet} & 23.20 & 0.917 &0.120 \\
			%		\hline
			DeepUPE~\cite{DUPE} & 23.24   & 0.893     & 0.158 \\
			%		\hline
			CURL~\cite{2019CURL}        &  24.20   & 0.880     & 0.108 \\
			%		\hline
			Deeplpf~\cite{DeepLPF}   &  24.48   & 0.887   & 0.103  \\
			%		\hline
			CSRNet~\cite{CSRNet}   & 25.06 & 0.935  &  0.090    \\
			%		\hline
			3DLUT~\cite{3DLUT}    &  24.92   & 0.934     & 0.093  \\		
			%		\hline
			LLFlow~\cite{LLFlow}  &  23.07  & 0.918    & 0.126 \\
			%		\hline
			BasicEnhancer~\cite{StarEnhancer}    &  25.46   & \textbf{0.948}     & 0.083  \\
			%		\hline
			Ours   &  \textbf{25.57} & {0.944} & \textbf{0.079} \\
			\bottomrule
		\end{tabular}
	\end{center}
	%	\vspace{-6mm}
\end{table}

% ------------------------------------------------------------------------------
\subsection{Learning Objective\label{sec:method_training}}
\label{sec:method_loss}

With the normalizing flow architecture~\cite{NICE,RealNVP,GLOW}, we are able to describe the probability distribution of the image tone style.
Following \cite{cFlow}, we can formulate the image tone style distribution via the change-of-variables formula, \ie,
\begin{equation}
	\label{density}
	\begin{split}
		\mathit{p}_\mathbf{s}(\mathbf{s};\mathbf{X},\theta_\mathcal{F})=\mathit{p}_\mathbf{z}(\mathcal{F}(\mathbf{s};\mathbf{X},\theta_\mathcal{F}))\left|\mathrm{det}\left(\frac{\partial \mathcal{F}}{\partial \mathbf{s}}\right)\right|,
	\end{split}
\end{equation}
where $J_\mathcal{F}(\mathbf{s}) = \frac{\partial \mathcal{F}}{\partial \mathbf{s}} $ denotes the Jacobian of $\mathcal{F}$ at $\mathbf{s}$.
Then, the normalizing flow can be trained by maximum likelihood estimation, which is usually implemented equivalently through minimizing the negative log-likelihood, \ie,
\begin{equation}
	\begin{split}
		\mathcal{L}_\mathit{NLL} = -\log{\mathit{p}_{\mathbf{z}}}(\mathcal{F}(\mathbf{s}; \mathbf{X},\theta_\mathcal{F})) - \log|\mathrm{det}\mathit{J}_\mathcal{F}(\mathbf{s})|.
	\end{split}
	\label{eqn:loss_NLL}
\end{equation}

We apply the progressive reconstruction loss $\mathcal{L}_\mathit{Retouching}$ on the progressive reconstructed results. $\mathcal{L}_\mathit{Retouching}$ updates the parameters of Style Encoder and the conditional RetouchNet concurrently, \ie,
\begin{equation}
	\mathcal{L}_\mathit{Retouch} = \sum_{t=1}^{n} \left \| \hat{\mathbf{Y}}^{(\mathit{t})}-\mathbf{Y} \right \|_1,
	\label{eqn:loss_retouching}
\end{equation}
where $\hat{\mathbf{Y}}^{(\mathit{t})}$ is the retouching results at each step. We set $n=3$ in our experiments.
In summary, the learning objective is,
\begin{equation}
	\mathcal{L} = \mathcal{L}_\mathit{Retouch} + \lambda\mathcal{L}_\mathit{NLL},
\end{equation}
where $\lambda$ is the hyperparameter for balancing the loss terms, and we empirically set $\lambda=1$ in our experiments.

\begin{table}[t]
	\caption{Quantitative comparisons on PPR10K dataset. The PPR10K-a/b/c denotes the portrait photo retouched by three experts. The results are tested on $360$p images.}
	\vspace{-1em}
	\setlength\tabcolsep{1.2pt}
	\renewcommand\arraystretch{1.2}%.25}
	\footnotesize
	\label{ppr10k}
	\begin{center}
		\begin{tabular}{ccccccc}
			%			\toprule
			\toprule
			{Method}&{Dataset}& {$PSNR \uparrow$} & {$\triangle E_{ab} \downarrow$} & {$PSNR^{HC} \uparrow$} &{$\triangle E_{ab}^{HC} \downarrow$} \\
			\midrule
			HDRNet~\cite{HDRNet} &PPR10K-a&23.93&8.70&27.21&5.65 \\
			CSRNet~\cite{CSRNet}  &PPR10K-a&22.72&9.75&25.90&6.33 \\
			3D LUT~\cite{3DLUT} &PPR10K-a&25.64&\textbf{6.97}&{28.89}&4.53 \\
			%			3D LUT+HRP~\cite{PPR10K}   &PPR10K-a&{25.99}&{6.76}&28.29&{4.38}\\
			Ours  &PPR10K-a&\textbf{25.72}& 6.98& \textbf{28.94}& \textbf{4.53} \\			
			\midrule
			HDRNet~\cite{HDRNet} &PPR10K-b&23.96&8.84&27.21&5.74\\
			CSRNet~\cite{CSRNet}  &PPR10K-b&23.76&8.77&27.01&5.68 \\
			3D LUT~\cite{3DLUT} &PPR10K-b&24.70&7.71&27.99&4.99 \\
			%			3D LUT+HRP~\cite{PPR10K}   &PPR10K-b&{25.06}&{7.51}&{28.36}&{4.85} \\
			Ours  &PPR10K-b&\textbf{24.82}& \textbf{7.70} & \textbf{28.04} & \textbf{4.98}  \\			
			\midrule
			HDRNet~\cite{HDRNet} &PPR10K-c&24.08&8.87&27.32&5.76 \\
			CSRNet~\cite{CSRNet}  &PPR10K-c&23.17&9.45&26.47&6.12 \\
			3D LUT~\cite{3DLUT} &PPR10K-c&25.18&\textbf{7.58}&28.49&\textbf{4.92} \\
			%			3D LUT+HRP~\cite{PPR10K}  &PPR10K-c&{25.46}&{7.43}&{28.80}&{4.82} \\
			Ours &PPR10K-c& \textbf{25.27} &{7.66}  & \textbf{28.51} & {4.97}  \\
			\bottomrule
			%			79, Eab: 8.0151, psnr_hc: 28.1227, Eab_hc: 5.2189, GLC:65/
		\end{tabular}
	\end{center}
\end{table}

\begin{figure*}[h]
    \small
    \centering
    \renewcommand\tabcolsep{2pt}
    \includegraphics[width=.95\linewidth]{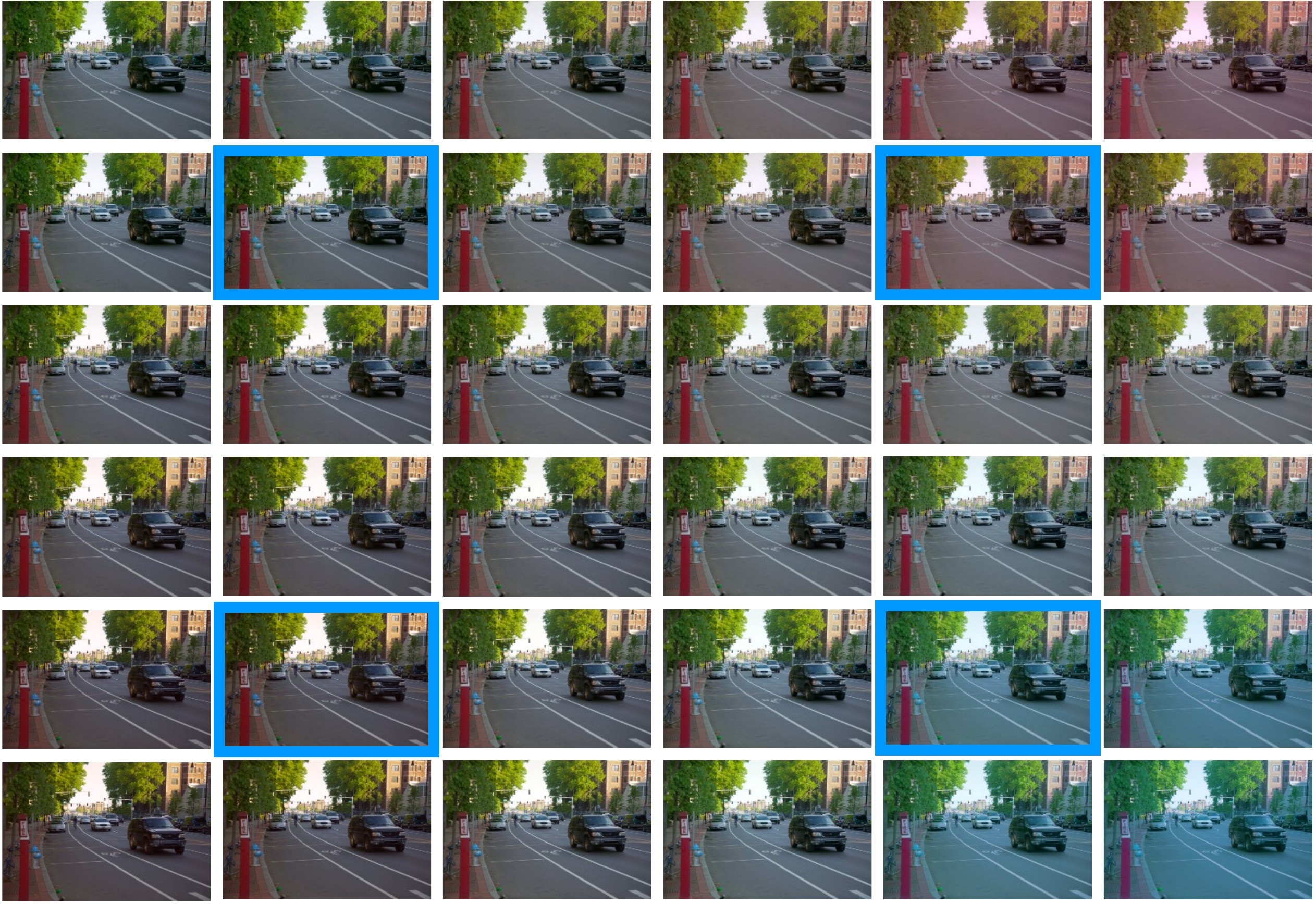} 
    \caption{Latent space interpolation/extrapolation. The images in blue boxes denote four different retouching results, while others are generated by interpolation/extrapolation in the latent space.}
    \label{fig:inter}
    \vspace{-1em}
\end{figure*}

\section{Experiments}

\begin{table} 
	\small
	\renewcommand\arraystretch{1.2}%.25}
	\caption{Quantitative comparison on MIT-Adobe FiveK dataset with diverse retouching results. The competing methods include StarEnhancer, CSRNet, and 3DLUT.}
	\label{tab:diverse}
	%	\vspace{-6mm}\multirow{2}*{Method}&\multirow{2}*{Dataset}& \multicolumn{2}{c}{$PSNR \uparrow$} <{\centering}p{1.0cm}
	\begin{center}
		\begin{tabular}{ccc}
			\toprule
			{\# of clustering centers} 	 & $\mathit{K}=2$  & $\mathit{K}=3$  \\
			\midrule
			Competing Methods    &  26.63 / 26.57 / 26.46& {27.24} \\	
			Ours     & \textbf{27.04} & \textbf{27.94} \\
			\hline
		\end{tabular}
	\end{center}
	%	\vspace{-6mm}
\end{table}

\subsection{Implementation Details}

\vspace{0.5em}
\noindent\textbf{Datasets}
Our experiments are performed on the MIT-Adobe FiveK dataset~\cite{fiveK} and the PPR10K dataset~\cite{PPR10K}.
The MIT-Adobe FiveK dataset~\cite{fiveK} contains 5,000 RAW images, and for each RAW image, we use five expert-retouched reference images (A/B/C/D/E) and two auto-retouching results (X/Y).
We follow the pre-processing pipeline and split setting in \cite{StarEnhancer}, where the first 4,500 samples are for training, and the rest 500 samples are for testing.
%
% To evaluate the flexibility of generating diverse unseen tone styles, we take the auto-retouching results (X/Y) as the test set, which are not involved in the training process.
% Finally, we calculate the PSNR~\cite{PSNR}, SSIM~\cite{SSIM}, and LPIPS~\cite{LPIPS} metrics for comparison with other methods.
%
The PPR10K dataset~\cite{PPR10K} contains 11,161 portrait photos with 1,681 groups, and each RAW photo is processed by three experts (a/b/c) with professional experience.
Following \cite{PPR10K}, the training set in our experiments contains 8,875 photos with 1,356 identities, while the remaining 2,286 photos with 325 identities form the test set. 
We employ the 360p setting in \cite{PPR10K}, where {the shorter edges of the portrait photos are resized to 360 pixels}.
% We evaluate our method quantitatively using PSNR~\cite{PSNR}, CIELAB color difference~\cite{DelatE} ($\Delta E_{ab}$). $*^{HC}$ denotes the metrics on human-centered areas. 

\vspace{0.5em}
\noindent\textbf{Training Settings and Evaluation Metrics}
During training, we feed 16 images into the model in each iteration, which are randomly cropped into $256\times256$ patches. The proposed framework is optimized by Adam algorithm~\cite{Adam} with a learning rate of $4\times10^{-5}$ and $(\beta_1, \beta_2)=(0.9, 0.999)$. 
After training, we follow the previous works~\cite{StarEnhancer,PPR10K}, which calculate the PSNR~\cite{PSNR}, SSIM~\cite{SSIM}, and LPIPS~\cite{LPIPS} metrics for the MIT-Adobe FiveK dataset~\cite{fiveK}, and compare the PSNR~\cite{PSNR} and CIELAB color difference\footnote{\url{https://en.wikipedia.org/wiki/Color_difference}} ($\Delta E_{ab}$) for the PPR10K dataset~\cite{PPR10K}.
All experiments are conducted using the PyTorch framework \cite{PyTorch} with an Nvidia GeForce RTX 2080Ti GPU.

\subsection{Comparison with State-of-the-art Methods}
\label{sec:exp_comparison}
Since our goal is to learn diverse tone styles for image retouching, the proposed model is designed to produce multiple retouching results for a specific input image via sampling diverse latent code $\mathbf{z}$'s.
However, traditional image retouching methods generate only one result for each input or style.
In order to compare with state-of-the-art image retouching methods in a fair scheme, we propose a method to obtain a latent code $\mathbf{\bar{z}}^\ast$ which represents the overall image retouching style of an expert.
Specifically, once the training is finished, we can obtain a style representation $\mathbf{z}_\mathit{i}$ for each training image with the Style Encoder $\mathit{E}$ (see \cref{eqn:encoder}).
Then, the average latent code $\mathbf{\bar{z}}$ is obtained by,
\begin{equation}
    \mathbf{\bar{z}}=\frac{1}{\mathit{N}}\sum_{\mathit{i}=1}^{\mathit{N}}{\mathbf{z}_\mathit{i}},
    \label{eqn:average_z}
\end{equation}
which is utilized to initialize the following optimization,
\begin{equation}
    \mathbf{\bar{z}}^\ast=\arg\min_{\mathbf{z}}\sum_\mathit{i=1}^\mathit{N}\|\mathit{G}(\mathbf{X}_\mathit{i}, \mathcal{F}^{-1}(\mathbf{z}; \mathbf{X}_\mathit{i},\theta_\mathcal{F});\theta_\mathit{G})-\mathbf{Y}_\mathit{i}\|_1,
    \label{eqn:latent_code_optimization}
\end{equation}
where $\mathit{N}$ is the number of training samples.
In this way, we can obtain the results via
\begin{equation}
    \mathbf{\hat{Y}}_\mathit{i}=\mathit{G}(\mathbf{X}_\mathit{i}, \mathcal{F}^{-1}(\mathbf{\bar{z}}^\ast; \mathbf{X}_\mathit{i},\theta_\mathcal{F});\theta_\mathit{G}).
    \label{eqn:generate_results}
\end{equation}

\cref{tab:quantitative results} shows the quantitative comparison against the state-of-the-art methods on the expert-C subset of MIT-Adobe FiveK dataset~\cite{fiveK}, including White-Box~\cite{hu2018exposure}, Distort-and-Recover~\cite{park2018distort}, DUPE~\cite{DUPE}, Pix2Pix~\cite{pix2pix}, Deeplpf~\cite{DeepLPF}, CSRNet~\cite{CSRNet}, 3D LUT~\cite{3DLUT}, and Starenhancer~\cite{StarEnhancer}, and the visual results are given in \cref{fig:SOTA fiveK}.
Besides, we also compare with state-of-the-art methods on PPR10K dataset~\cite{PPR10K}, including HDRNet~\cite{HDRNet}, CSRNet~\cite{CSRNet}, and 3D LUT~\cite{3DLUT}, and the qualitative and quantitative results are given in \cref{fig:sota PPR10K} and \cref{ppr10k}.

Note that all competing models are trained and evaluated under the same dataset configuration.
According to the results, one can see that our method can achieve comparable or superior performance against the state-of-the-art methods in describing the overall retouching style of an expert, even though it is trained to learn the image tone style distribution of an expert with intrinsic diversity.

\subsection{Exploring the Intrinsic Diversity of an Expert}
\label{sec:exp_diversity}
In \cref{sec:exp_comparison}, the power of our method is greatly limited since only one retouching style can be generated for comparison.
However, our model is trained to describe an image retouching style distribution, which captures the intrinsic diversity of an expert, rather than a specific style.
Therefore, the performance of our model can be further enhanced by generating multiple candidate retouching results for each input image.
In particular, instead of averaging all latent codes like \cref{eqn:average_z}, we can obtain multiple (\eg, $\mathit{K}$) representative latent codes ($\{\mathbf{\bar{z}}_\mathit{j}\}_{\mathit{j}=1}^\mathit{K}$) by clustering with the k-means algorithm~\cite{kmeans}, which are supposed to better describe the image retouching style distribution of the expert.
When generating final results, we select the best one among the $\mathit{K}$ results, \ie,
\begin{equation}
\begin{split}
    \mathbf{\hat{Y}}_\mathit{i}&=\arg\max_{\mathbf{\hat{Y}}_\mathit{i,j}}\mathrm{PSNR}(\mathbf{\hat{Y}}_\mathit{i,j},\mathbf{Y}_\mathit{i})\end{split}
\label{eqn:diverse_results_1}
\end{equation}
with
\begin{equation}
\begin{split}
    \mathbf{\hat{Y}}_\mathit{i,j}&=\mathit{G}(\mathbf{X}_\mathit{i}, \mathcal{F}^{-1}(\mathbf{\bar{z}}_\mathit{j}; \mathbf{X}_\mathit{i},\theta_\mathcal{F});\theta_\mathit{G}).
\end{split}
\label{eqn:diverse_results_2}
\end{equation}

In \cref{tab:diverse}, we show the results when there are 2 or 3 clustering centers.
Accordingly, we also select three best-performed competing methods (\ie, CSRNet~\cite{CSRNet}, 3D LUT~\cite{3DLUT}, and StarEnhancer~\cite{StarEnhancer}) for comparison.
Specifically, when there are $\mathit{K}$ clustering centers for our method, we take $\mathit{K}$ methods from them, and process each test sample with the $\mathit{K}$ methods.
Analogous to \cref{eqn:diverse_results_1}, the retouching result with the highest PSNR among the $\mathit{K}$ results is utilized.
Note that all conditions of selecting two from three methods are listed in \cref{tab:diverse}. According to \cref{tab:diverse}, our method can outperform all the competing methods, showing the superior ability to describe the retouching style distribution. Our method can generate diverse results easily without consuming memory to contain additional models.

\begin{figure*}[t]
    \centering
    \includegraphics[width=\linewidth]{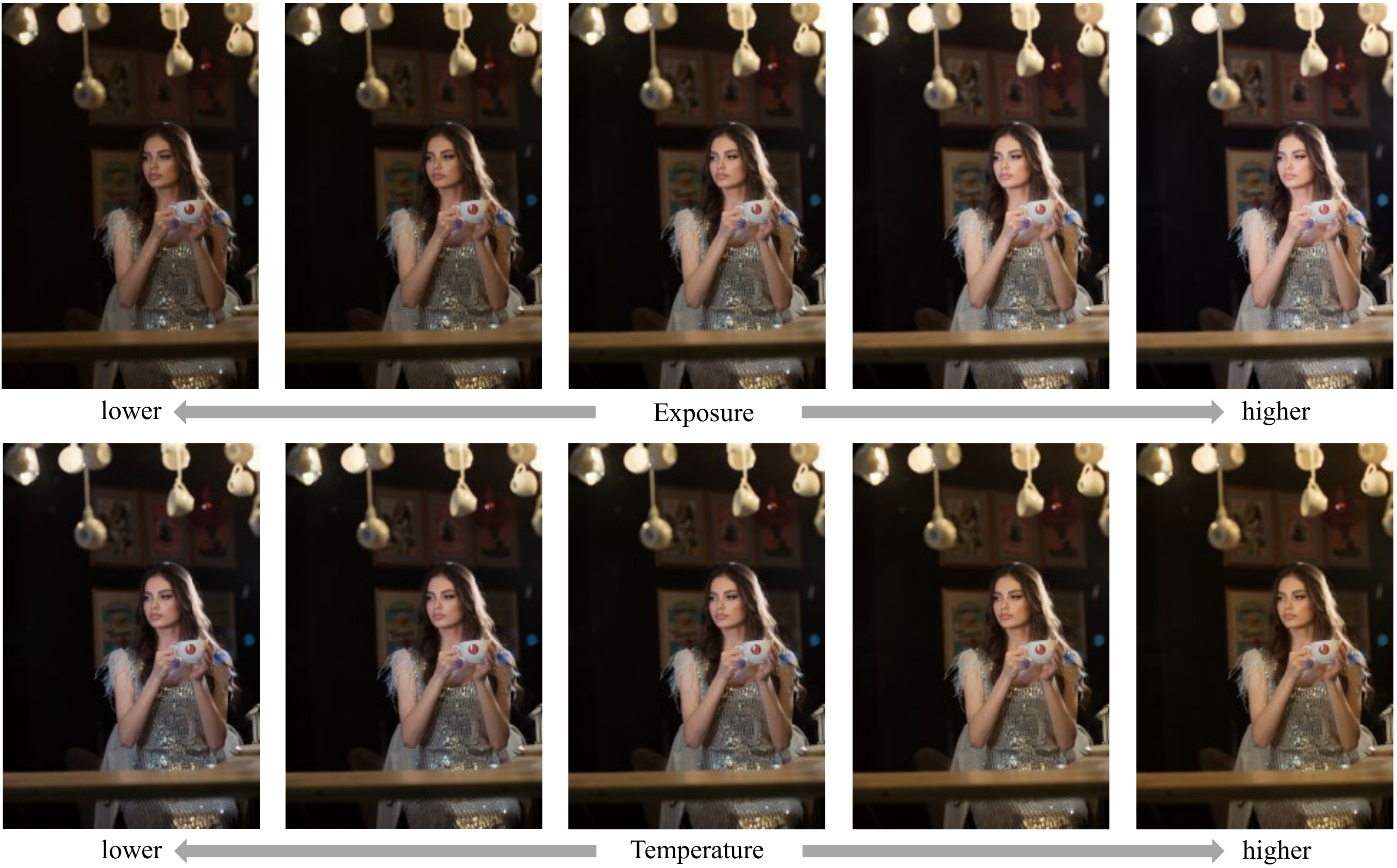}\\
    \caption{Visual retouching results based on adjusting a single dimension of the latent variable. For example, the retouching results show differences in image exposure and temperature, respectively.}
    \label{fig:controllable}
\end{figure*}

\begin{table*}
	\small
	\caption{Quantitative comparison on diverse style retouching. Note that StarEnhancer~\cite{StarEnhancer} is trained with all subsets (A-Y), Ours is trained with only expert-C subset, and Ours$^\dagger$ is trained with five subsets (A-E). Unseen styles during training are highlighted in gray.}
	\label{tab:quantitative multiple}
	\begin{center}
		\begin{tabular}{ccccccccc}
			%			\toprule
			\toprule
			Method &  \footnotesize{A} & \footnotesize{B}  & \footnotesize{C} &\footnotesize{D} &\footnotesize{E} &\footnotesize{X}&\footnotesize{Y}&\footnotesize{Average}\\
			\midrule
			StarEnhancer~\cite{StarEnhancer} & 19.63 & 25.18 &25.29 &22.79& \textbf{24.07}& \textbf{28.82} &23.85 &24.23 \\
			Ours & \cellcolor[RGB]{239,239,239}{20.31} & \cellcolor[RGB]{239,239,239}25.29 & \textbf{25.57} & \cellcolor[RGB]{239,239,239}22.34 & \cellcolor[RGB]{239,239,239}23.29 & \cellcolor[RGB]{239,239,239}26.90 & \cellcolor[RGB]{239,239,239}22.96 & 23.81 \\		
			Ours$^\dagger$ & \textbf{20.74} &\textbf{26.17}& {25.36}  & \textbf{22.89} &{23.66} & \cellcolor[RGB]{239,239,239}28.15 & \cellcolor[RGB]{239,239,239}\textbf{24.24} & \textbf{24.46}\\
			\bottomrule
		\end{tabular}
	\end{center}
\end{table*}

\subsection{Unseen Retouching Style Generation}
The experiments in \cref{sec:exp_comparison,sec:exp_diversity} are still focused on approximating the retouching style of the training set.
As shown in \cref{fig:diverseretouching}, our method can produce diverse image retouching results by sampling multiple $\mathbf{z}$'s from the Gaussian distribution, where consistency is observed among different samples processed with the same $\mathbf{z}$.
In other words, our method provides stable and robust unseen retouching styles rather than random ones.
For further verification, we exploit more subsets of the MIT-Adobe FiveK dataset~\cite{fiveK}, \ie, five subsets processed by expert-A/B/C/D/E and two subsets automatically generated by Adobe Lightroom (denoted by X and Y).
In particular, for generating the style of a particular subset, we reuse the formulation of \cref{eqn:average_z,eqn:latent_code_optimization,eqn:generate_results}, and the only difference is the source of $\mathbf{z}_\mathit{i}$.
As shown in \cref{tab:quantitative multiple}, we provide the results of our models on generating retouching styles of the seven styles.
Note that \textit{Ours} means the model trained with only expert-C subset, while \textit{Ours}$^\dagger$ means the model trained with five expert-retouched subsets (A/B/C/D/E).
We also provide the results of StarEnhancer~\cite{StarEnhancer} for comparison.
One can see that even trained with only expert-C subset, \textit{Ours} can achieve decent performance on other styles.
When trained with more subsets, \textit{Ours}$^\dagger$ achieves the best performance.

\begin{table}[t]
	\caption{Quantitative results of our model and its variants on predicting expert retouching style. $n$ denotes the number of flow steps in TSFlow.}
	\vspace{-1em}
	\footnotesize
	\label{tab:ablationpredict}
	\begin{center}
		\begin{tabular}{cccc}
			\toprule
			%\multirow{2}*{Measure}& \multicolumn
			Method&$PSNR\uparrow$&$SSIM\uparrow$&$LPIPS\downarrow$ \\
			\midrule
			TSFlow (\textit{nocond.})  &  22.62 & 0.888  &  0.115 \\ 
			TSFlow (step=4)  & 25.12  & 0.939  &  0.083 \\ 
			TSFlow (step=12)  &  25.59 & 0.958  &  0.083 \\
						\midrule
			Ours w/o ISE         & 24.62 & 0.888  & 0.115  \\ 
			%				\hline
			Ours w/o PSC          & 25.23 & 0.942  &  0.082  \\ 
						\midrule
			%				\hline
			Ours         & \textbf{25.57} & \textbf{0.944} & \textbf{0.079}  \\
			\bottomrule
		\end{tabular}
	\end{center}
\end{table}

\subsection{Ablation Study \label{ablation}}

\begin{figure*}[t]
	\centering
    \begin{minipage}{0.13\linewidth}
		\includegraphics[width=\linewidth]{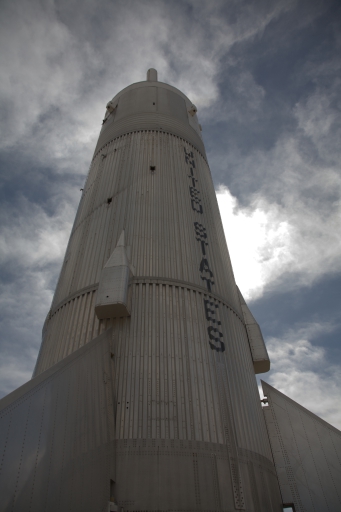}\\
		\vspace{-3mm}
		\includegraphics[width=\linewidth]{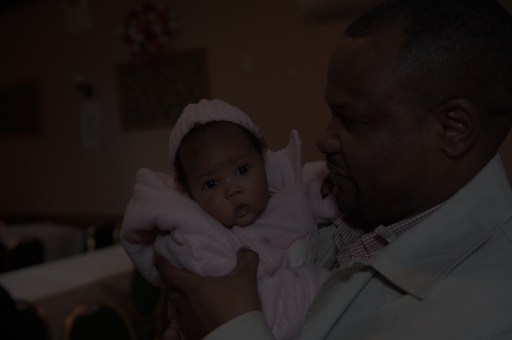}\\
		\vspace{-3mm}
		\includegraphics[width=\linewidth]{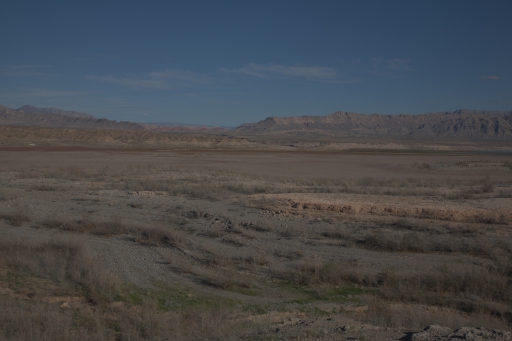}
		\caption*{\footnotesize Input}
	\end{minipage}
	\begin{minipage}{0.13\linewidth}
		\includegraphics[width=\linewidth]{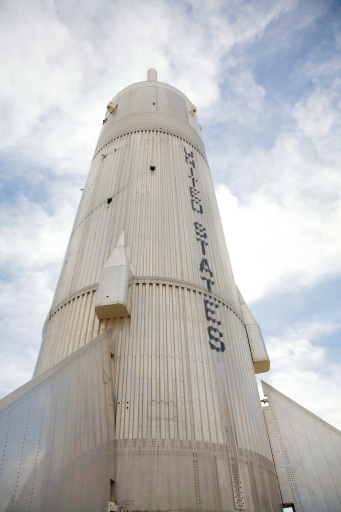}\\
		\vspace{-3mm}
		\includegraphics[width=\linewidth]{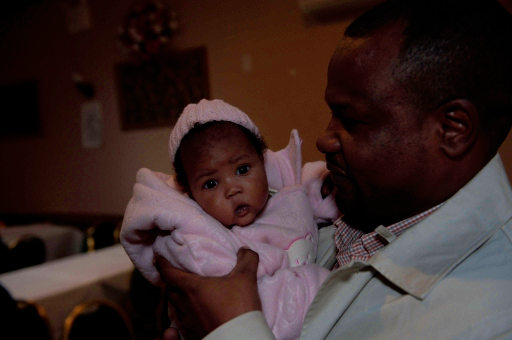}\\
		\vspace{-3mm}
		\includegraphics[width=\linewidth]{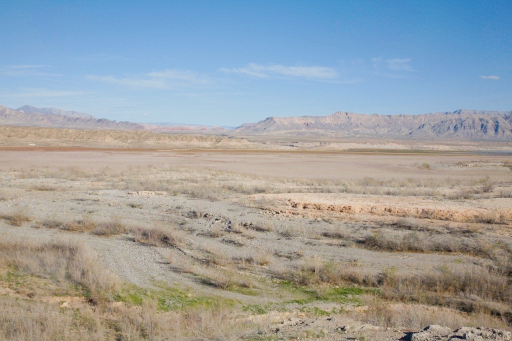}
		\caption*{\footnotesize TSFlow (\textit{nocond.})}
	\end{minipage}
	\begin{minipage}{0.13\linewidth}
		\includegraphics[width=\linewidth]{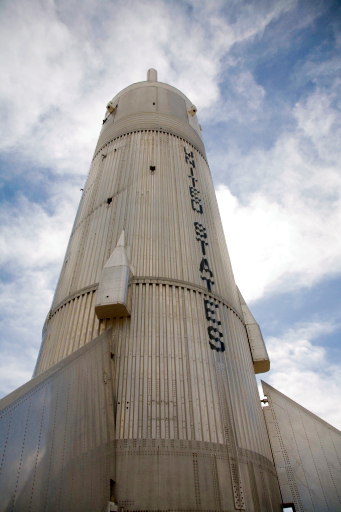}\\
		\vspace{-3mm}
		\includegraphics[width=\linewidth]{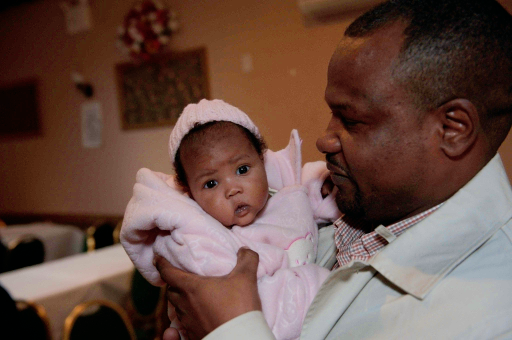}\\
		\vspace{-3mm}
		\includegraphics[width=\linewidth]{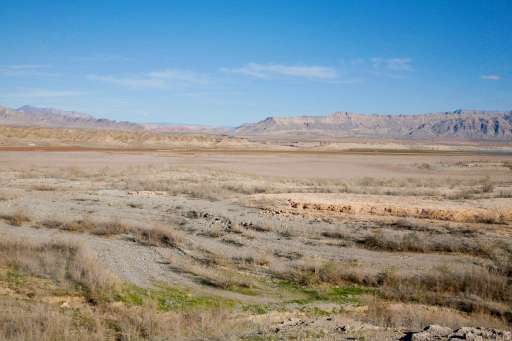}
		\caption*{\footnotesize TSFlow (step=4)}
	\end{minipage}
	\begin{minipage}{0.13\linewidth}
		\includegraphics[width=\linewidth]{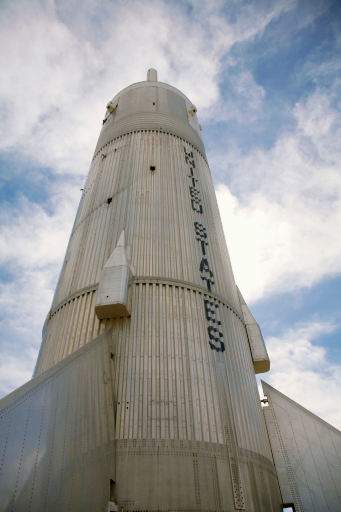}\\
		\vspace{-3mm}
		\includegraphics[width=\linewidth]{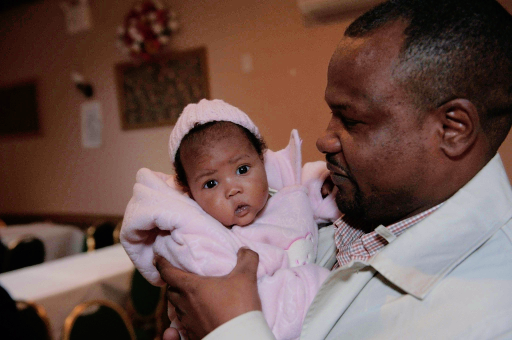}\\
		\vspace{-3mm}
		\includegraphics[width=\linewidth]{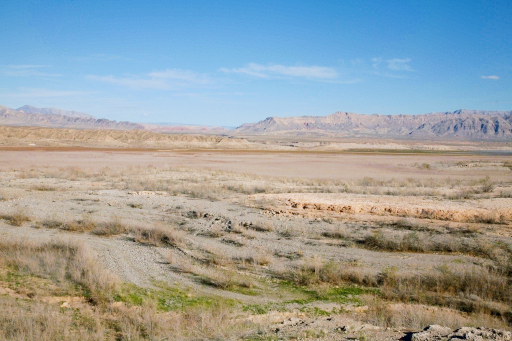}
		\caption*{\footnotesize Ours w/o ISE}
	\end{minipage}
	\begin{minipage}{0.13\linewidth}
		\includegraphics[width=\linewidth]{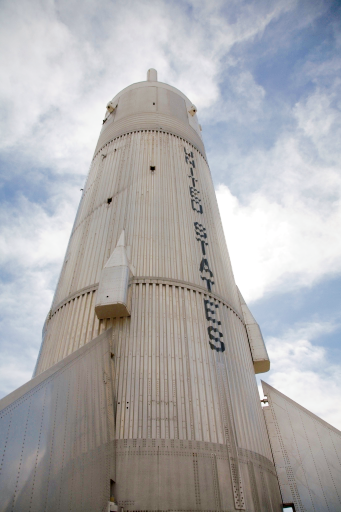}\\
		\vspace{-3mm}
		\includegraphics[width=\linewidth]{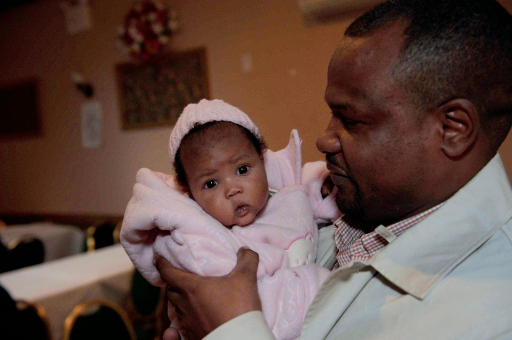}\\
		\vspace{-3mm}
		\includegraphics[width=\linewidth]{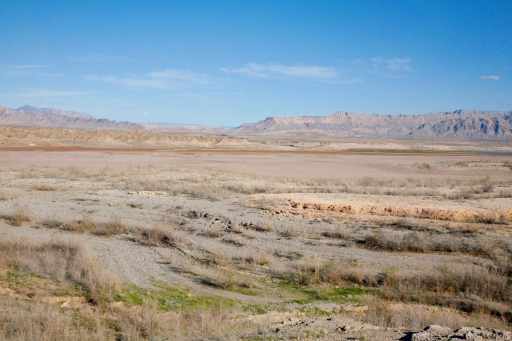}
		\caption*{\footnotesize Ours w/o PSC}
	\end{minipage}
	\begin{minipage}{0.13\linewidth}
		\includegraphics[width=\linewidth]{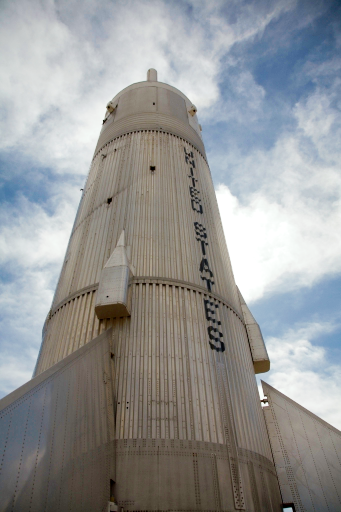}\\
		\vspace{-3mm}
		\includegraphics[width=\linewidth]{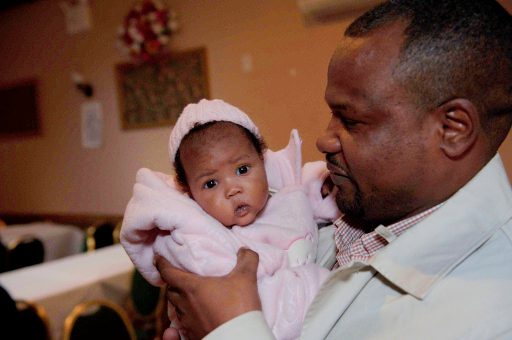}\\
		\vspace{-3mm}
		\includegraphics[width=\linewidth]{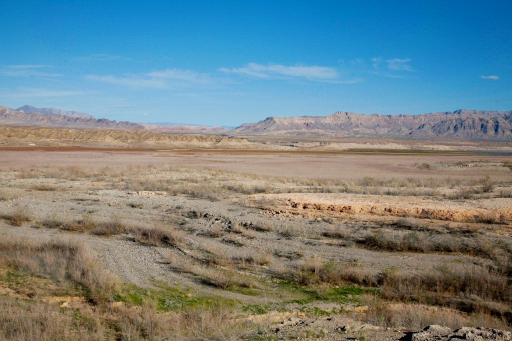}
		\caption*{\footnotesize Ours}
	\end{minipage}
	\begin{minipage}{0.13\linewidth}
		\includegraphics[width=\linewidth]{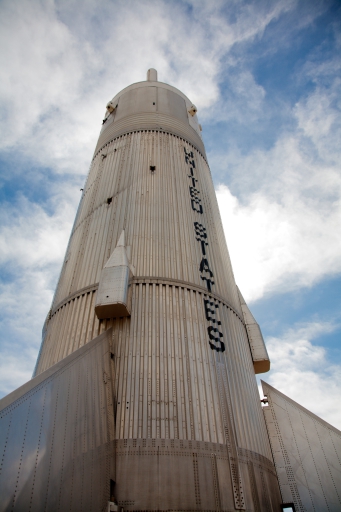}\\
		\vspace{-3mm}
		\includegraphics[width=\linewidth]{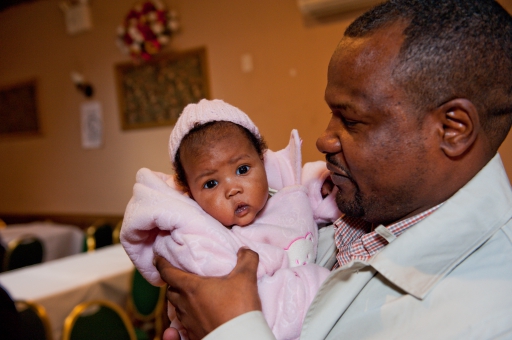}\\
		\vspace{-3mm}
		\includegraphics[width=\linewidth]{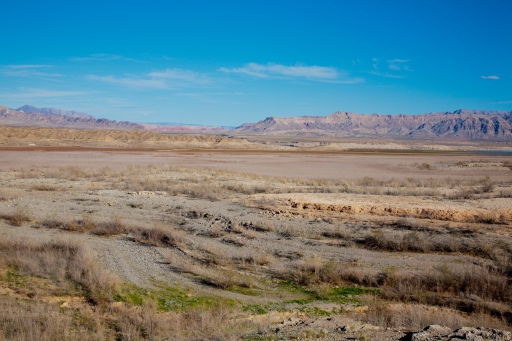}
		\caption*{\footnotesize Reference}
	\end{minipage}
	\caption{Qualitative results of our model and its variants on predicting retouching style.}
	\label{fig:ablation expert}
\end{figure*}

\vspace{0.5em}
\noindent\textbf{Latent Space Exploration.}
With the TSFlow module, the retouching style distribution is mapped into a latent space (\ie, a simple Gaussian distribution).
To explore the property of the latent space, we conduct two experiments.
(i)~Latent code interpolation: we select four latent codes which generate different retouching styles, and obtain more latent codes via interpolation/extrapolation.
As shown in \cref{fig:inter}, the interpolated/extrapolated latent codes can generate retouching styles with regular patterns.
(ii)~Latent code modification: thanks to the disentanglement ability of the normalizing flow framework, we further explore the influence of particular dimensions of the latent code $\mathbf{z}$.
As shown in \cref{fig:controllable}, by modifying different dimensions of $\mathbf{z}$, we can observe diverse retouching results with varying exposure and color temperature.
% Thus each dimension of latent variables $z$ exists interpretability. 
%The reason is TSFlow learns the bi-direction mapping between the isotropic normal distribution and the image tone style distritbuion.
%
In other words, our method can provide an interactive adjustment for image retouching by operations in the latent space, which further satisfies the users' aesthetic preferences more conveniently.

\vspace{0.5em}
\noindent\textbf{Image Tone Style Representation Quality.} 
One of the key differences between our method and previous ones is that we regard the retouching style of an expert as a distribution, and describe the image tone style of each retouched image separately.
For improving the image tone style representation quality, image-specific style extraction and progressive style correction are utilized in our training scheme.
To verify the effectiveness of these components, we conduct experiments with two variants.
The first discards the image-specific style extraction design (denoted by ``w/o ISE''), while the second detaches the progressive style correction mechanism (denoted by ``w/o PSC'').
As shown in \cref{tab:ablationpredict}, without ISE, the PSNR drops by nearly 1 dB, and performance degradation is also observed without PSC.
Please refer to \cref{fig:ablation expert} for visual comparison.

%In our proposed framework, the high quality of the image tone style guarantee that TSFlow learn  reliable target distribution. 
%
% To investigate the effectiveness of image tone style quality on our method, we trained two variant models.
% The first variant of Ours w/o ISE (Image-Specific Style Encoder) replaces the Style Encoder takes the input images solely and estimate the tone styles. 
%
% Fig.~\ref{fig:ablation res} shows that the estimated tone style achieve less performance on reconstructing expert retouched images. Thus it has low quality and cannot describe the distribution of expert retouching style. When trained on these tone style vectors, TSFlow get non-idealy supervision and achieve low performance on predicting expert retouching style (Fig.~\ref{fig:ablation expert}). 
% %
% The second variant is Our w/o PSC (Progressive Style Correction). Compared with Ours, there exist error on the reconstructed images. Thus the extracted image tone style still carry less retouching representation.
% %
% Therefore our proposed ISE and PSC can produce high quality image tone style vectors, which benefit on providing more reliable supervision for TSFlow. Quantitative comparison are shown in Table.~\ref{ablationreconstruct} and Table.~\ref{ablationpredict}.

\vspace{0.5em}
\noindent \textbf{The Structure of TSFlow.}
We further conduct experiments to evaluate the design of the TSFlow structure.
As shown in \cref{tab:ablationpredict}, TSFlow (\textit{uncond.}) means the unconditional variant of our TSFlow, where the input image $\mathbf{X}$ is no longer deployed as the condition information for the TSFlow module.
Such modification leads to severe image quality degradation, showing that the retouching style of an expert is altered according to the image content, and our TSFlow can well capture such knowledge.
We also show the influence of the number of flow steps.
The experiments on the 4, 8, and 12 flow steps show that 8 flow steps can achieve a trade-off between the computation complexity and performance, which is taken as the final design of our TSFlow module.
The visual comparison is also given in \cref{fig:ablation expert}.

\section{Conclusion} 
Existing image retouching methods perform deterministic retouching results, which are inflexible to meet the subjective preferences of users.
In this paper, we propose a novel normalizing flow-based framework for generating diverse retouching images.
To solve the spatial disharmony effect caused by existing normalizing flow-based methods, we propose an image tone style normalizing flow (TSFlow) module to learn the conditional distribution of the image tone styles, and disentangle the style representation with the image content.
An image-specific style encoder and a progressive style correction mechanism are proposed to extract high-quality image tone style representations from the expert-retouched images.
Extensive experiments show that our proposed method outperforms the state-of-the-art methods, and can achieve diverse image retouching styles to satisfy different user preferences.

\section*{Acknowledgments}
This work was supported in part by the National Natural Science Foundation of China (NSFC) under Grant No. U19A2073 and No. 62006064.

{\small
	\bibliographystyle{IEEEtran}
	\bibliography{egbib}
}

%\newpage

%needed around the contents of the optional argument to biography to prevent
%the LaTeX parser from getting confused when it sees the complicated
%$\backslash${\tt{includegraphics}} command within an optional argument. (You can create
%your own custom macro containing the $\backslash${\tt{includegraphics}} command to make things
%simpler here.)
%\bf{If you include a photo:}\vspace{-33pt}
%\begin{IEEEbiography}[{\includegraphics[width=1in,height=1.25in,clip,keepaspectratio]{fig1}}]{Michael Shell}
%Use $\backslash${\tt{begin\{IEEEbiography\}}} and then for the 1st argument use $\backslash${\tt{includegraphics}} to declare and link the author photo.
%Use the author name as the 3rd argument followed by the biography text.
%\end{IEEEbiography}

\vfill

\end{document}